\documentclass[letterpaper, 10 pt, conference]{ieeeconf}  
\pdfoutput=1

\IEEEoverridecommandlockouts                              

\usepackage{times}
\usepackage{graphicx} 
\usepackage{graphicx} 
\usepackage{subfigure} 
\usepackage{multicol}
\usepackage[small,it]{caption}
\usepackage{verbatim}
\usepackage{amsmath}
\usepackage{wrapfig,multirow}
\usepackage{array}
\usepackage{relsize}
\usepackage{amssymb,amsmath,mathtools}
\usepackage{url}

\usepackage{rotating}

\usepackage{graphicx}
\usepackage{subfigure}
\usepackage{multirow}
\usepackage{multicol}
\usepackage{times}
\let\labelindent\relax
\usepackage{enumitem}
\usepackage{xcolor}
\usepackage{verbatim}
\usepackage{hyperref}
\usepackage{array}
\usepackage{relsize}
\usepackage{amssymb,amsmath}
\usepackage{url}
\usepackage{rotating}
\usepackage{wrapfig,booktabs,blindtext}
\usepackage{color}

\usepackage{color}

\graphicspath{{fig/}}

 \belowdisplayskip \abovedisplayskip
\renewcommand{\baselinestretch}{0.97}

 \newlength\savedwidth

\newlength{\sectionReduceTop}
\newlength{\sectionReduceBot}
\newlength{\subsectionReduceTop}
\newlength{\subsectionReduceBot}
\newlength{\abstractReduceTop}
\newlength{\abstractReduceBot}
\newlength{\captionReduceTop}
\newlength{\captionReduceBot}
\newlength{\subsubsectionReduceTop}
\newlength{\subsubsectionReduceBot}

\newlength{\horSkip}
\newlength{\verSkip}

\newlength{\figureHeight}
\title{\LARGE \bf PlanIt: A Crowdsourcing Approach for Learning to Plan Paths
\\from Large Scale Preference Feedback}

\author{Ashesh Jain, Debarghya Das, Jayesh K. Gupta and Ashutosh Saxena
\thanks{A. Jain, D. Das, J. K. Gupta and A. Saxena are with the Department of Computer Science,
Cornell University, USA. {\tt ashesh@cs.cornell.edu, dd367@cornell.edu,
mail@rejuvyesh.com,
asaxena@cs.cornell.edu}}%
}

\begin{document}

\maketitle


\begin{abstract}
We consider the problem of learning user preferences over robot trajectories for environments rich in objects and humans. This is challenging because the criterion defining a good trajectory varies with users, tasks and interactions in the environment. We represent trajectory preferences using a cost function that the robot learns and uses it to generate good trajectories in new environments.  We design a crowdsourcing system - PlanIt, where non-expert users label segments of the robot's trajectory. PlanIt allows us to collect a large amount of user feedback, and using the weak and noisy labels from PlanIt we learn the parameters of our model. We test our approach on 122 different environments for robotic navigation and manipulation tasks. Our extensive experiments show that the learned cost function  generates preferred trajectories in human environments. Our crowdsourcing system is publicly available for the visualization of the learned costs and for providing preference feedback: \url{http://planit.cs.cornell.edu}
\end{abstract}

\section{Introduction}
\vspace*{\sectionReduceBot}
\vspace*{\sectionReduceBot}
\indent    One key problem robots face in performing tasks in human environments
is identifying trajectories desirable to the users. 
In this work we present a crowdsourcing system PlanIt that learns user
preferences by taking their feedback over the Internet.
In previous works, user preferences are usually encoded as a cost over trajectories,
and then optimized using planners such as RRT*~\cite{Karaman10}, CHOMP~\cite{Chomp},
TrajOpt~\cite{Schulman13}. However, most of these works optimize expert-designed
cost functions based on different geometric and safety
criteria~\cite{Sisbot07,Sisbot12,Mainprice11}. While satisfying safety criteria
is necessary, they alone ignore 
the contextual interactions in human environments~\cite{Jain13b}.  We take a data 
driven approach and learn a context-rich cost over the trajectories
from the preferences shown by \textit{non-expert} users.

\begin{figure}[t]
\centering
\includegraphics[width=1\linewidth,natwidth=1434,natheight=813]{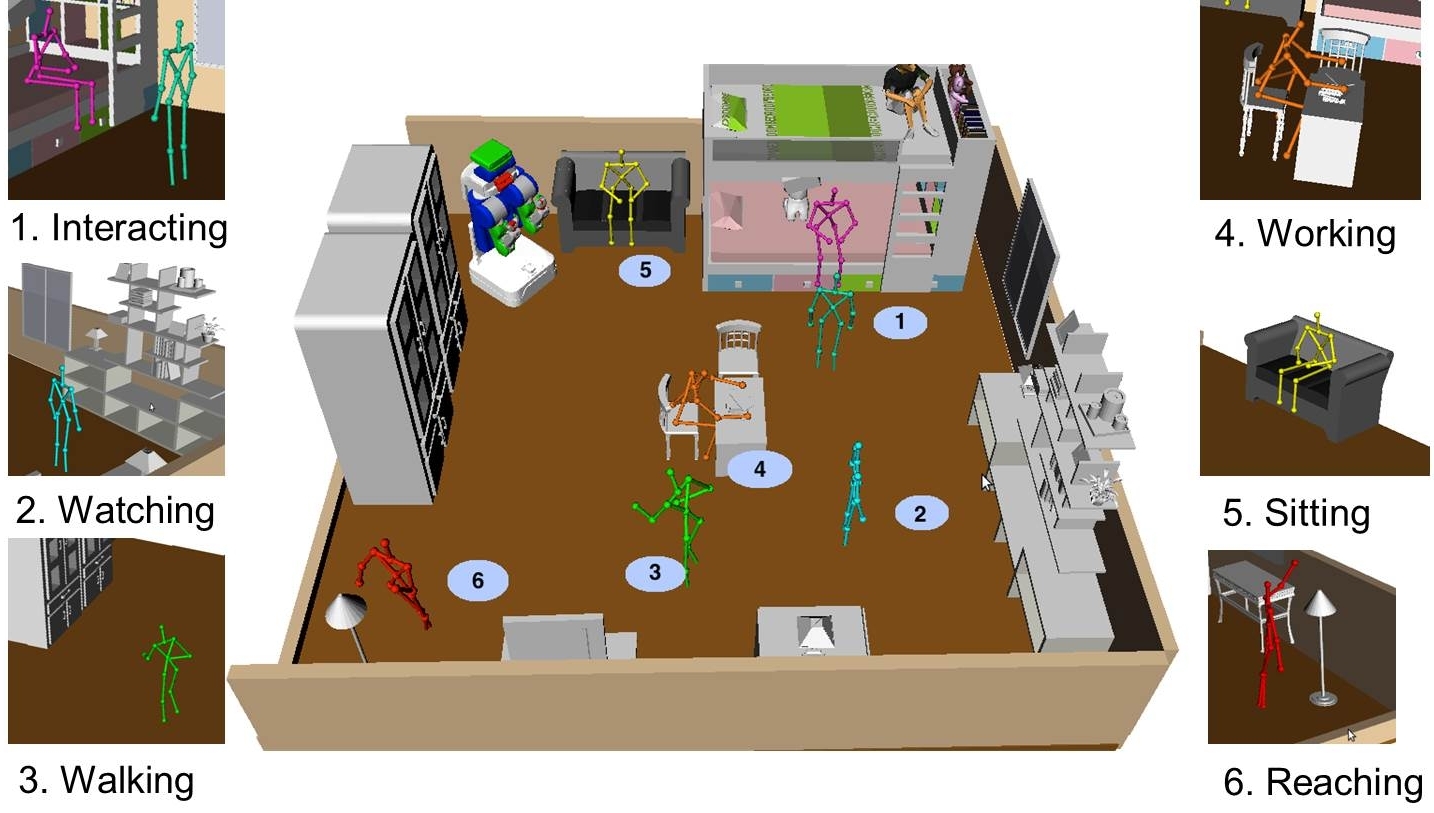}
\vspace*{2\captionReduceTop}
\caption{\textbf{Various human activities with the objects in the environment} affect how a robot should 
navigate in the environment. The figure shows an environment with multiple human activities: (1)  two humans
\textit{interacting}, (2) \textit{watching}, (3) \textit{walking}, (4) \textit{working}, (5) \textit{sitting}, and (6) \textit{reaching} for a lamp.
We learn a spatial distribution for each activity, and use it to build a cost map (aka
planning affordance map) for the complete environment. Using the cost map, the robot
plans a preferred trajectory in the environment.}
\vspace*{\captionReduceBot}
\vspace*{\captionReduceBot}
\label{fig:activities}
\end{figure}

In this work we model user preferences arising during human activities. Humans constantly engage in 
activities with their surroundings -- watching TV or listening to music, etc. -- during which 
they prefer minimal interruption from external agents that share their environment.
For example, a robot that blocks the view of a human watching TV is not a
 desirable social agent. \textit{How can a
robot learn such preferences and context?}
This problem is further challenging because human environments are unstructured, and as shown in Fig.~\ref{fig:activities} an environment can have multiple  human activities happening simultaneously. Therefore generalizing the learned model to new environments is a key challenge.

We formulate the problem as learning to ground each human activity to a spatial
distribution signifying regions crucial to the activity. We refer to these
spatial distributions as \textit{planning
  affordances}\footnote{Gibson~\cite{Gibson86} defined object affordances  as
  possible actions that an agent can perform in an environment.}  and
parameterize the cost function  using these distributions. 
Our affordance representation is different by relating to the object's
functionality, unlike previous works which have an object centric view.
The commonly studied discrete representation of affordances~\cite{Sahin07,Montesano08,Koppula13b,Katz13} are of limited use in planning trajectories. For example, a TV has a \textit{watchable} affordance and undergoes a \textit{watching} activity, however these labels themselves are not informative enough to convey to the robot that it should not move between the user and the TV. The grounded representation we propose in this work is more useful for planning tasks than the discrete representations.

To generalize well across diverse environments 
we develop a crowdsourcing web-service \texttt{PlanIt} to collect large-scale
preference data. 
On PlanIt we show short videos (mostly $<$ 15 sec) to non-expert users of the robot navigating
in context-rich environments with humans performing activities.
As feedback users label segments of the videos as good, bad or neutral.
While previous methods of eliciting feedback required expensive expert demonstrations in limited environments,
PlanIt is usable by \textit{non-expert} users and scales to a large number of environments.
This simplicity comes at the cost of weak and noisy feedback.
We present a generative model of the preference data obtained.

We evaluate our approach on a total of 122 bedroom and living room environments.
We use OpenRave~\cite{Diankov10} to generate 
trajectories in these environments and upload them to PlanIt database for user feedback.
We quantitatively evaluate our learned model and compare it to previous
works on human-aware planning.
Further, we validate our model on the PR2 robot to navigate in human environments.
The results show that our learned model generalizes well to the environments not seen before.

In the following sections, we formally
state the planning problem, give an overview of the PlanIt engine in
Section~\ref{sec:planit}, discuss the cost parametrization through affordance in
Section~\ref{sec:representation},
describe the learning algorithm in
Section~\ref{subsec:learning}, and show the experimental evaluation in
Section~\ref{sec:experiment}.

\section{Related Work}
\vspace*{.0\sectionReduceBot}
\label{sec:related}
\textbf{Learning from demonstration (LfD).} One approach to learning preferences is to mimic an expert's demonstrations. Several works have built on this idea such as the autonomous helicopter flights~\cite{Abbeel10}, the ball-in-a-cup experiment~\cite{Kober11}, 
planning 2-D paths~\cite{Ratliff06}, etc. These
approaches are applicable in our setting. However, they are expensive in that
they require an expert to demonstrate the optimal trajectory. Such demonstrations 
are difficult to elicit on a large scale and over many environments. 
Instead we learn with preference data from non-expert users
across a wide variety of environments. 

\textbf{Planning from a cost function.} In many applications, the goal is to find a trajectory that optimizes a cost function. 
Several works build upon the sampling based planner RRT~\cite{Lavalle01,Karaman10} to optimize various cost heuristics~\cite{Ferguson06,Jaillet10}. Some
approaches introduce sampling bias~\cite{Leven03} to guide the planner. 
Alternative approaches include recent trajectory optimizers CHOMP~\cite{Chomp} and TrajOpt~\cite{Schulman13}.
We are complementary to these works in that we learn a cost function while the above approaches optimize cost functions.

\textbf{Modeling human motion for navigation path.} Sharing environment with humans requires robots to model
and predict human navigation patterns and generate socially compliant
paths~\cite{Bennewitz05,Kuderer12,Ziebart09}. 
Recent works~\cite{Koppula13,Mainprice13,Wang13} model human motion to
anticipate their actions for better human-robot collaboration. Instead we model
the spatial distribution of human activities and the preferences associated with
those activities.

\textbf{Affordances in robotics.} Many works in robotics have studied
affordances. Most of the works study affordance as cause-effect relations, i.e.
the effects of robot's actions on objects~\cite{Sahin07,Ugur11,Uyanik13,Montesano08}. 
We differ from these works in 
the representation of affordance and in
its application to planning user preferred trajectories. 
Further, we consider context-rich environments where humans interact with various 
objects, while such context was not important to previous works. Similar to Jiang et al.~\cite{Jiang12}, 
our affordances are also distributions, but they used them for scene arrangement while we use them for planning. 

\textbf{User preferences in path planning.} User preferences have been studied in human-robot
interaction literature. Sisbot et al.~\cite{Sisbot07,Sisbot07b} and Mainprice et al.~\cite{Mainprice11}
planned trajectories satisfying user specified preferences such as 
the distance of the robot from humans, 
visibility of the robot and human arm comfort. Dragan et al.~\cite{Dragan13a} used functional gradients~\cite{Chomp} 
to optimize for legibility of robot trajectories. We differ from these in that
we \textit{learn} the cost function capturing preferences arising during 
human-object interactions. Jain et al.~\cite{Jain13,Jain13b} learned a
context-rich cost via iterative feedback from non-expert users. 
Similarly, we also learn from the preference data of non-expert users. 
However, we use crowdsourcing like Chung et al.~\cite{Chung14} for eliciting user feedback which allows us to
learn from large amount of preference data. In experiments, we compare against
Jain's trajectory preference perceptron algorithm.  


\section{Context-aware planning problem}
\begin{figure}[t]
\centering
\includegraphics[width=.6\linewidth,natwidth=1232,natheight=907]{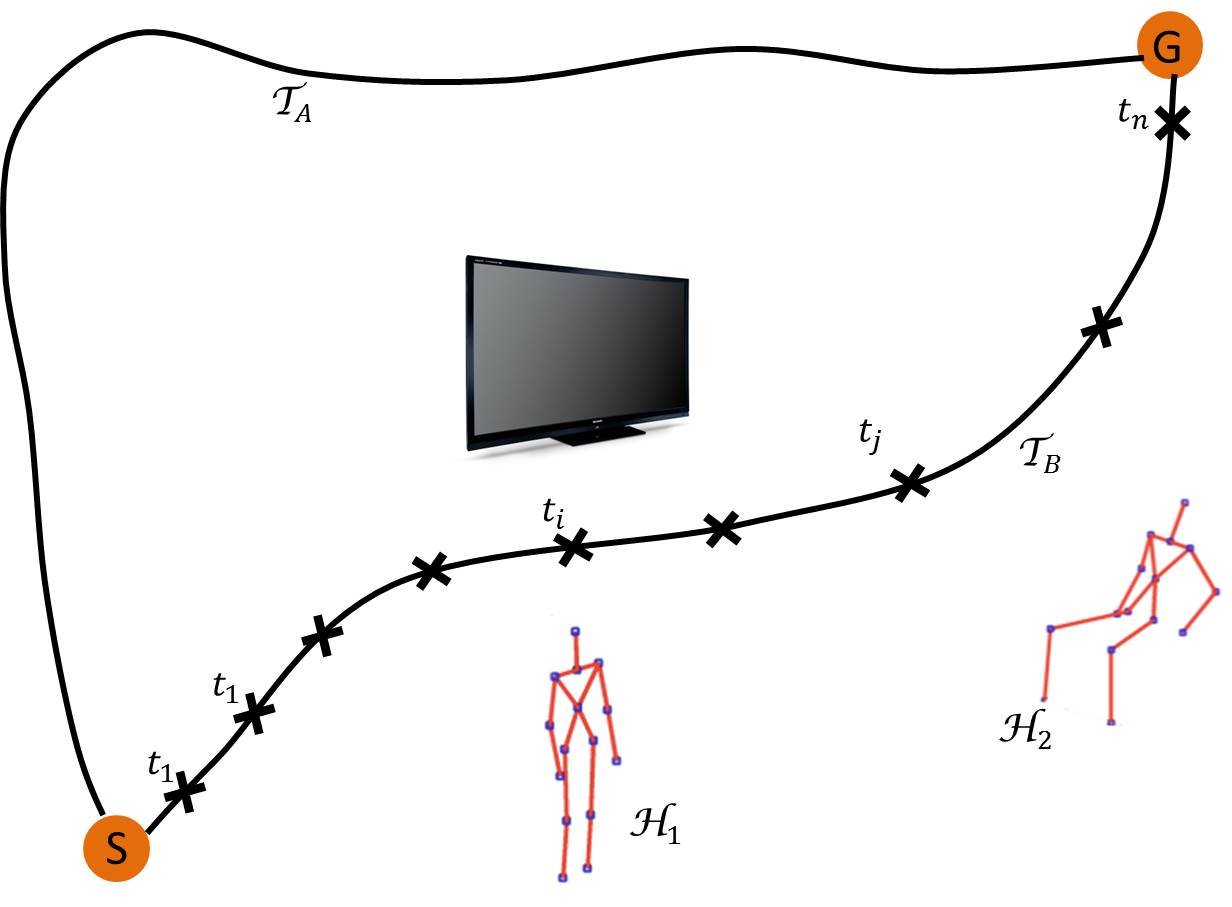}
\vspace*{\captionReduceTop}
\caption{\textbf{Preference-based Cost calculation of a trajectory.}   The trajectory $\mathcal{T}_A$ is preferred
over $\mathcal{T}_B$ because it does not interfere with the human activities.
The cost of a trajectory decomposes over the waypoints $t_i$, and the cost
depends on the location of the objects and humans associated with an activity.}
\vspace*{\captionReduceBot}
\vspace*{\captionReduceBot}
\label{fig:additive-cost}
\end{figure}

The planning problem we address is:  given a goal configuration $G$
and a context-rich environment $E$ (containing objects, humans and activities), 
the algorithm should output a desirable trajectory $\hat{\mathcal{T}}$.
We consider navigation trajectories  and represent them as a sequence of discrete
2D waypoints, i.e., $\mathcal{T} = \{t_1,\ldots,t_n\}$. Our model is easily 
extendable to higher dimensional manipulation trajectories, we demonstrate this in 
Section~\ref{sec:manip}.

In order to encode the user's desirability we use a 
positive cost function $\mathbf{\Psi}(\cdot)$ that maps trajectories to a scalar value. 
Trajectories with lower cost indicate greater desirability.
We denote the cost of trajectory 
$\mathcal{T}$ in environment $E$ as $\mathbf{\Psi}(\mathcal{T}|E)$ where
$\mathbf{\Psi}$ is defined as:
$$\mathbf{\Psi}|E: \mathcal{T} \longrightarrow \mathbb{R}$$

The context-rich environment $E$ comprises humans, objects and activities. Specifically,
it models the human-human and human-object interactions.
The robot's goal is to learn the spatial distribution of these interactions in order to plan good trajectories that minimally interrupt human activities. The key challenge here lies in designing an expressive cost
function that accurately reflects user preferences, captures the rich environment context, and can be learned from data.

Fig.~\ref{fig:additive-cost} illustrates how the cost of a trajectory is the cumulative effect of the environment at each waypoint.  We thus define a trajectory's cost as a product of the costs over each waypoint:
\begin{equation}
\mathbf{\Psi}(\mathcal{T}=\{t_1,..,t_n\}|E) = \prod_i \Psi_{a_i}(t_i|E)  \,\,
\label{eq:costfunction}
\end{equation}

In the above equation, $\Psi_{a_i}(t_i|E)$ is the cost of waypoint $t_i$ and its always positive.\footnote{Since the cost is always positive, the product of costs in Equation~\eqref{eq:costfunction} is equivalent to the sum of logarithmic cost.}
Because user preferences vary over activities, we learn a separate cost for each activity.
$\Psi_{a}(\cdot)$ denotes the cost associated with an activity $a\in E$. 
The robot navigating along  a trajectory often interferes with multiple human activities e.g., trajectory $\mathcal{T}_\mathcal{B}$
in Fig.~\ref{fig:additive-cost}.
Thus we associate with each waypoint $t_i$ an activity $a_i$ it interacts with, as illustrated in Eq.~\eqref{eq:costfunction}. 
\begin{figure}[t]
\centering
\includegraphics[width=\linewidth,natwidth=1509,natheight=664]{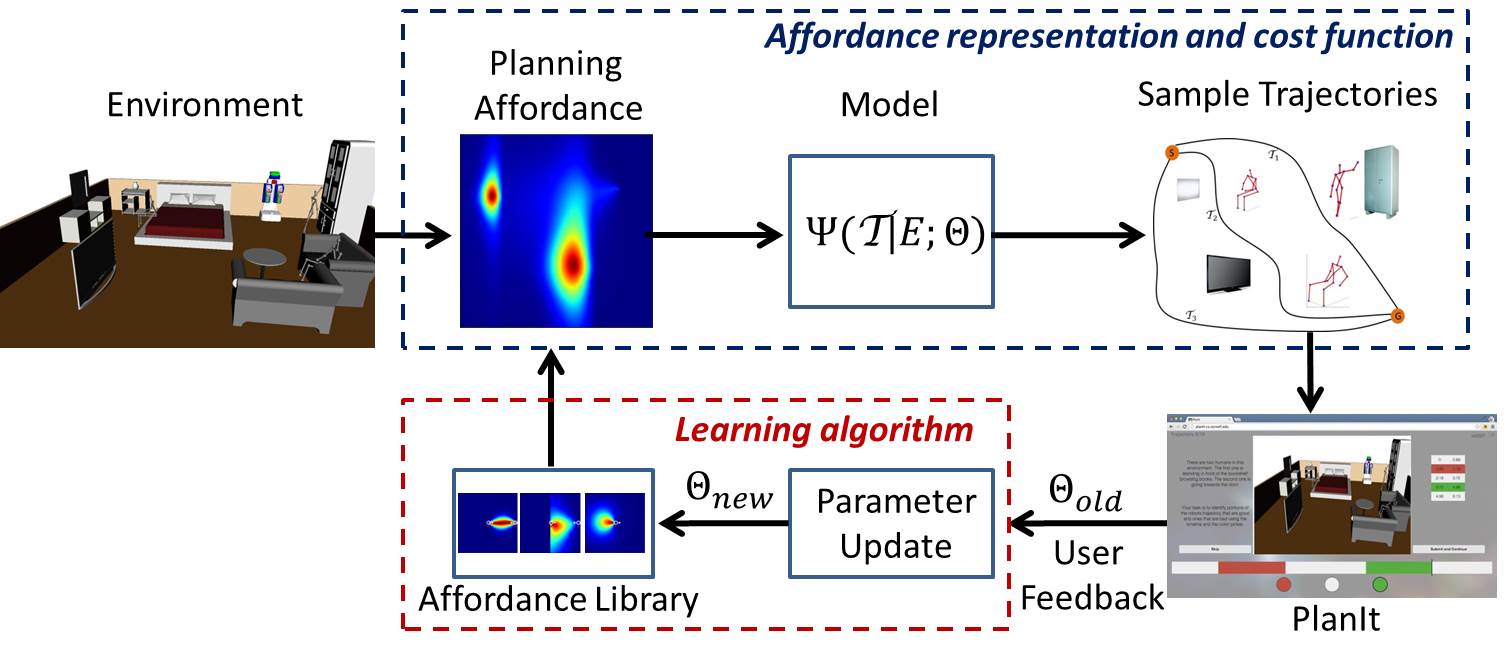}
\vspace*{2\captionReduceTop}
\caption{\textbf{An illustration of our PlanIt system.} Our learning system has
three main components (i) cost parameterization through affordance; 
(ii) The PlanIt engine for receiving user
preference feedback; and (iii) Learning algorithm. (Best viewed in color)}
\vspace*{\captionReduceBot}
\vspace*{\captionReduceBot}
\label{fig:system}
\end{figure}

The cost function changes with the activities happening in the environment.
As illustrated in Fig.~\ref{fig:additive-cost}, the robot prefers the trajectory $\mathcal{T}_A$ over the otherwise preferred shorter trajectory $\mathcal{T}_B$ because the latter interferes with human interactions (e.g., $H_2$ is watching TV).


\section{PlanIt: A crowdsourcing engine}
\vspace*{\sectionReduceBot}
\label{sec:planit}

\begin{figure}[t]
\centering
\includegraphics[width=\linewidth,natwidth=1370,natheight=765]{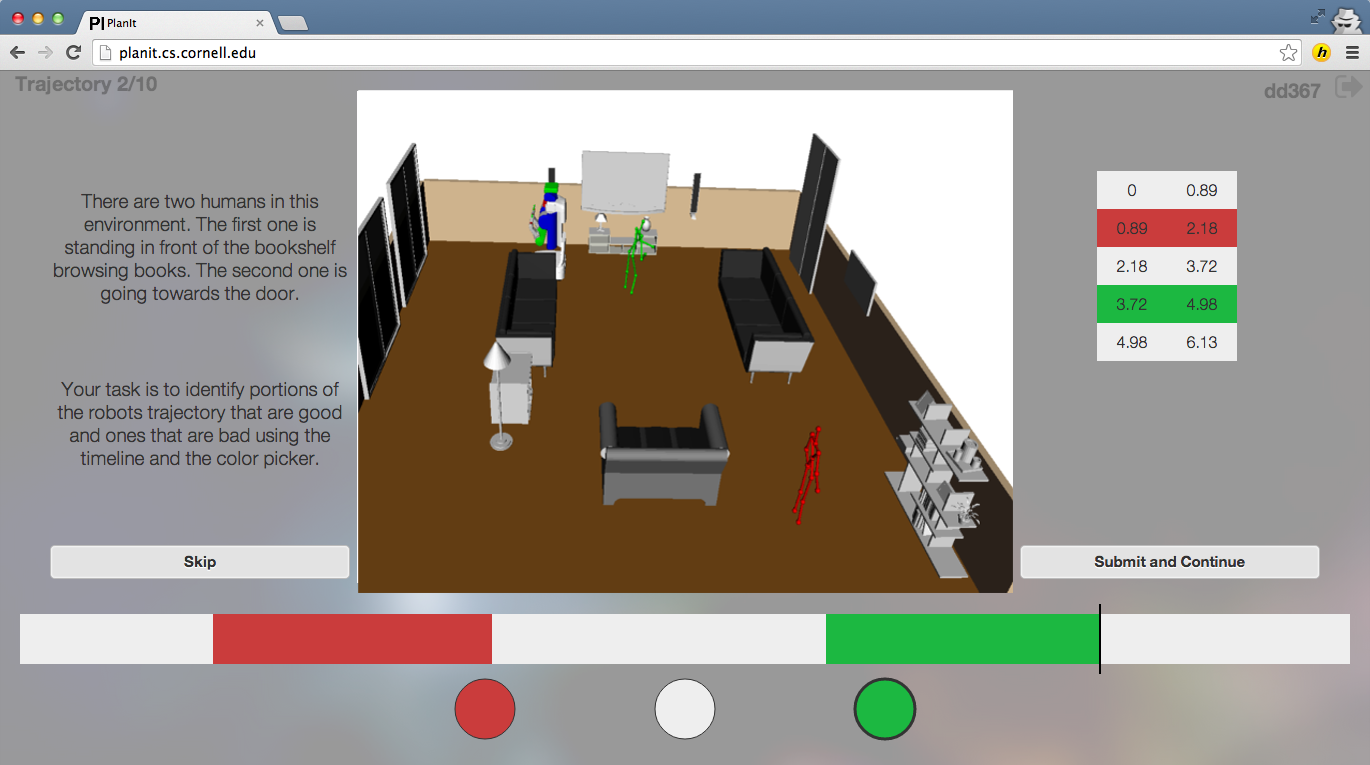}
\vspace*{2\captionReduceTop}
\caption{\textbf{PlanIt Interface.} Screenshot of the PlanIt
video labeling interface. The video shows a human walking towards the door 
while other human is browsing books, with text describing the environment on
left. 
As feedback the user labels the time interval  where the robot crosses the human
browsing books as red, and the interval where the robot carefully avoids the walking
human as green. (Best viewed in color)}
\vspace*{\captionReduceBot}
\vspace*{\captionReduceBot}
\label{fig:planit}
\end{figure}

Rich data along with principled learning algorithms have achieved much success
in robotics problems such as grasping~\cite{Curtis08,Miller04,Paolini13},
manipulation~\cite{Katz08}, trajectory
modeling~\cite{Vernaza12} etc. Inspired by such previous works, we design \emph{PlanIt}: a scalable approach 
for learning user preferences over robot trajectories across a wide-variety of
environments: \url{http://planit.cs.cornell.edu}

On PlanIt's webpage users 
watch videos of robot navigating in contextually-rich environments and reveal
their preferences
 by labeling video segments (Fig.~\ref{fig:planit}). 
We keep the process simple for users by providing three label choices
\{\textit{bad, neutral, good}\}. 
For example, the trajectory segments where the robot passes between a human and TV 
can be labeled as bad, and segments where it navigates in open space as
neutral. We now discuss three aspects of PlanIt.

\emph{A. Weak labels from PlanIt:}
In PlanIt's feedback process, users only label parts of a trajectory (i.e. sub-trajectory) as good, bad or neutral.
For the ease of usability and to reduce the
labeling effort, users only provide the labels and do not reveal the
(\textit{latent}) reason for the labels.
We capture the user's intention as a latent variable in the learning algorithm (discussed in Section~\ref{subsec:learning}).

The user feedback in PlanIt is in contrast to other learning-based approaches such as learning from the expert's demonstrations (LfD)~\cite{Abbeel10,Kober11,Ratliff06,Akgun12} or the co-active feedback~\cite{Jain13,Jain13b}. In both LfD and co-active learning approaches it is time consuming and expensive to collect the preference data on a robotic platform and across many environments. Hence these approaches learn using limited preference data from users.
On the other hand, PlanIt's main objective is to leverage the crowd and learn from the \textit{non-expert} users across a large number of  environments.

\emph{B. Generating robot trajectory videos:} 
We sample many trajectories (using RRT~\cite{Lavalle01}) for the PR2 robot in
human environments using OpenRAVE~\cite{Diankov10}. We video record these trajectories and add them to PlanIt's trajectory database. The users watch the short videos of the PR2 interacting with human activities, and reveal their preferences. We also ensure that trajectories in the database are diverse by following the ideas
presented in \cite{Berg10,Jain13}. As of now the PlantIt's database has 2500 trajectories over 122 environments. In Section~\ref{sec:experiment} we describe the data set.

\emph{C. Learning system:} In our learning system, illustrated in Fig.~\ref{fig:system}, the learned model improves as more preference data from users become available. We maintain an affordance library with spatial distributions for each human activity. When the robot observes an environment, it uses the distributions from the library and builds a planning affordance map (aka cost function) for the environment. 
The robot then samples trajectories from the cost function and presents them on
the PlanIt engine for feedback.
\section{Learning Algorithm}
We first discuss our parameterization of the cost function and then the
procedure for learning the model parameters.
\vspace*{\subsectionReduceTop}
\subsection{Cost Parameterization through Affordance}
\label{sec:representation}

In order to plan trajectories in human environments we model the human-object relationships.  These relationships are
called `object affordances'. In this work we model the affordances such that they are relevant to path planning and we refer to them as `planning affordances'.

Specifically, we learn the spatial distribution of the human-object interactions. For example, a TV has a \textit{watchable} affordance and therefore the space between the human and the TV is relevant for the \textit{watching} activity.  Since a \textit{watchable}   label  by itself is not informative enough to help in planning we ground it to a spatial distribution. Fig.~\ref{fig:angular-pref-1}(a) illustrates 
the learned spatial distribution when the human watches TV. 
Similarly, a chair is \textit{sittable} and \textit{moveable}, but when in use the space behind the chair is critical (because the human sitting on it might move back). 

\begin{figure}[t]
\centering
\begin{tabular}{@{\hskip -.1in}c@{\hskip -.2in}c@{\hskip -.1in}}
\subfigure[Interacting]{
\includegraphics[width=.5\linewidth,height=.33\linewidth, natwidth=576,natheight=432]{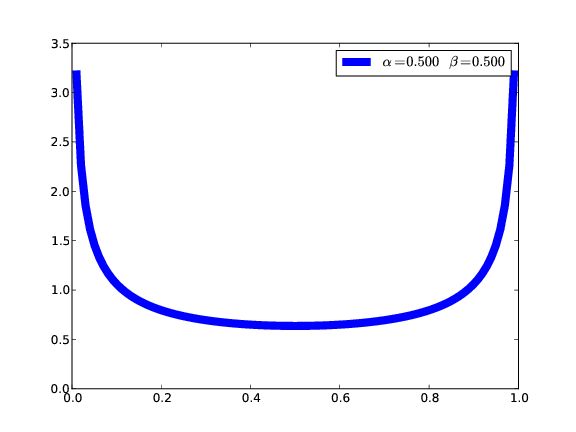}
}
&
\subfigure[Watching]{
  \includegraphics[width=.5\linewidth,height=.33\linewidth,natwidth=576,natheight=432]{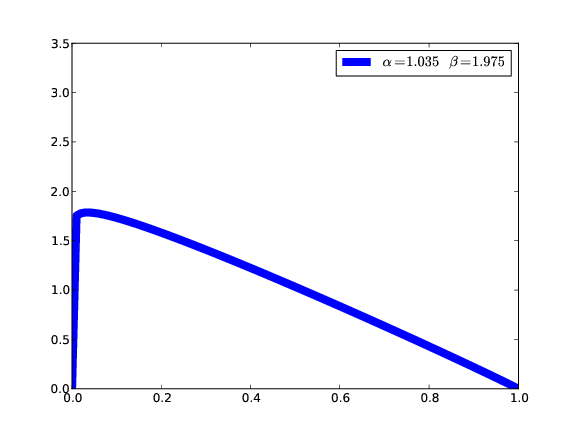}
}
\end{tabular}
\vspace*{\captionReduceTop}
\caption{\textbf{Result of learned edge preference.} Distance between human and
object is normalized to 1. Human is at 0 and object at 1. For interacting
activity, edge preference is symmetric between two humans, but for watching
activity humans do not prefer the robot passing very close to them.}
\vspace*{\captionReduceBot}
\label{fig:edge-pref}
\end{figure}

\begin{figure}[t]
\centering
\begin{tabular}{ccc}
\subfigure[Watching]{
\includegraphics[width=.25\linewidth,natwidth=1130,natheight=1101]{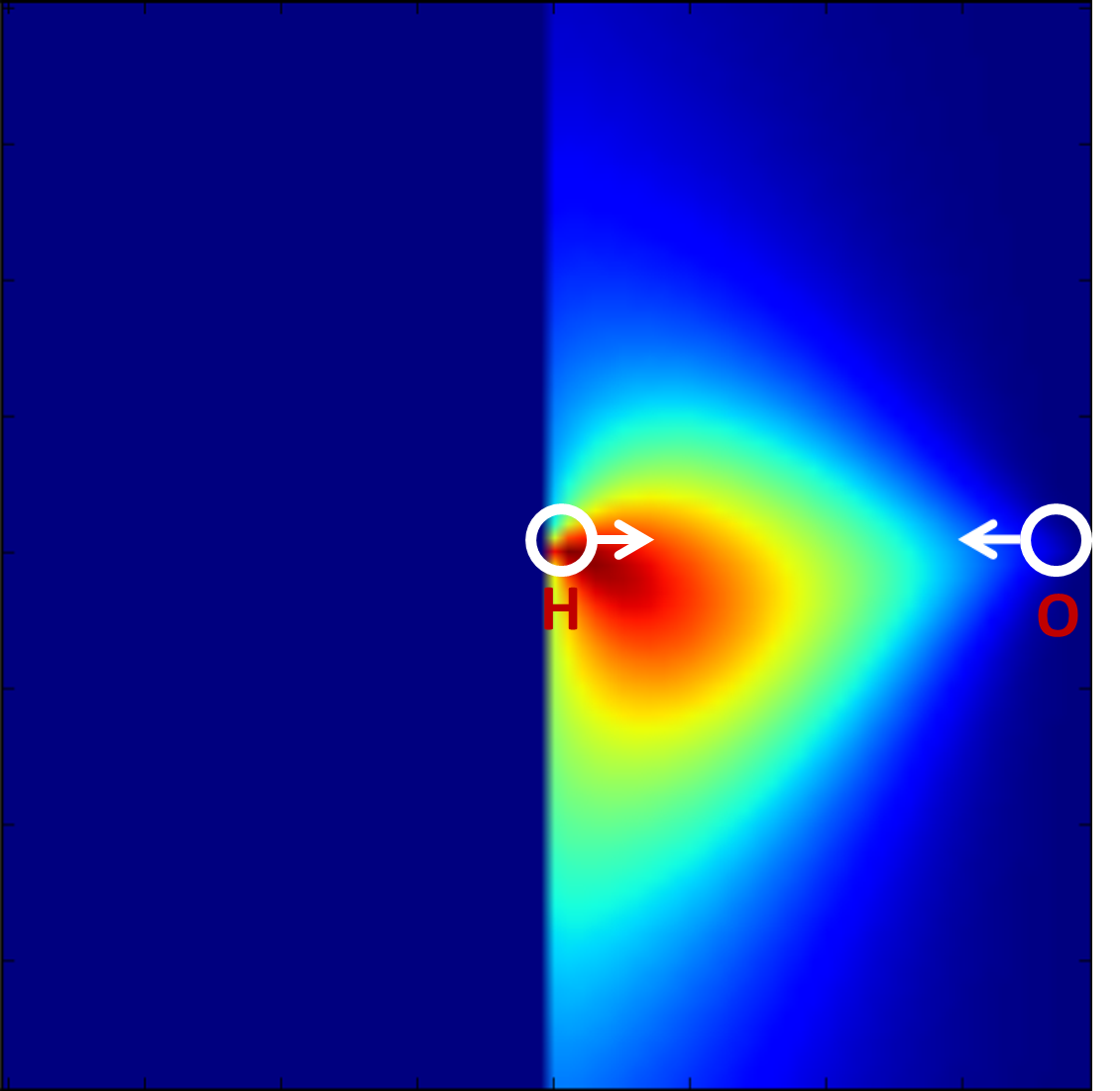}
}
&
\subfigure[Interacting]{
\includegraphics[width=.25\linewidth,natwidth=1131,natheight=1101]{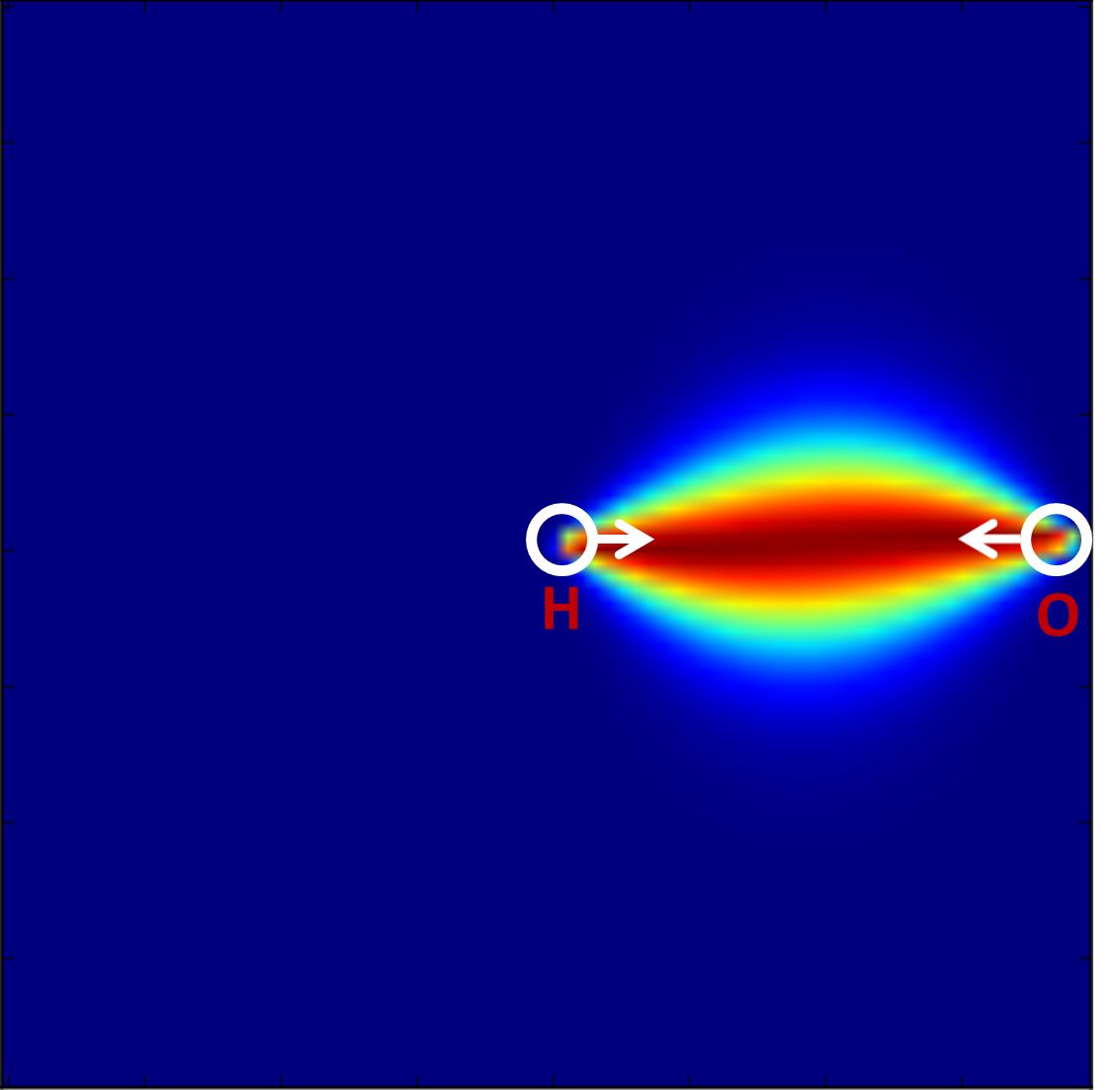}
}
&
\subfigure[Working]{
\includegraphics[width=.25\linewidth,natwidth=1104,natheight=1103]{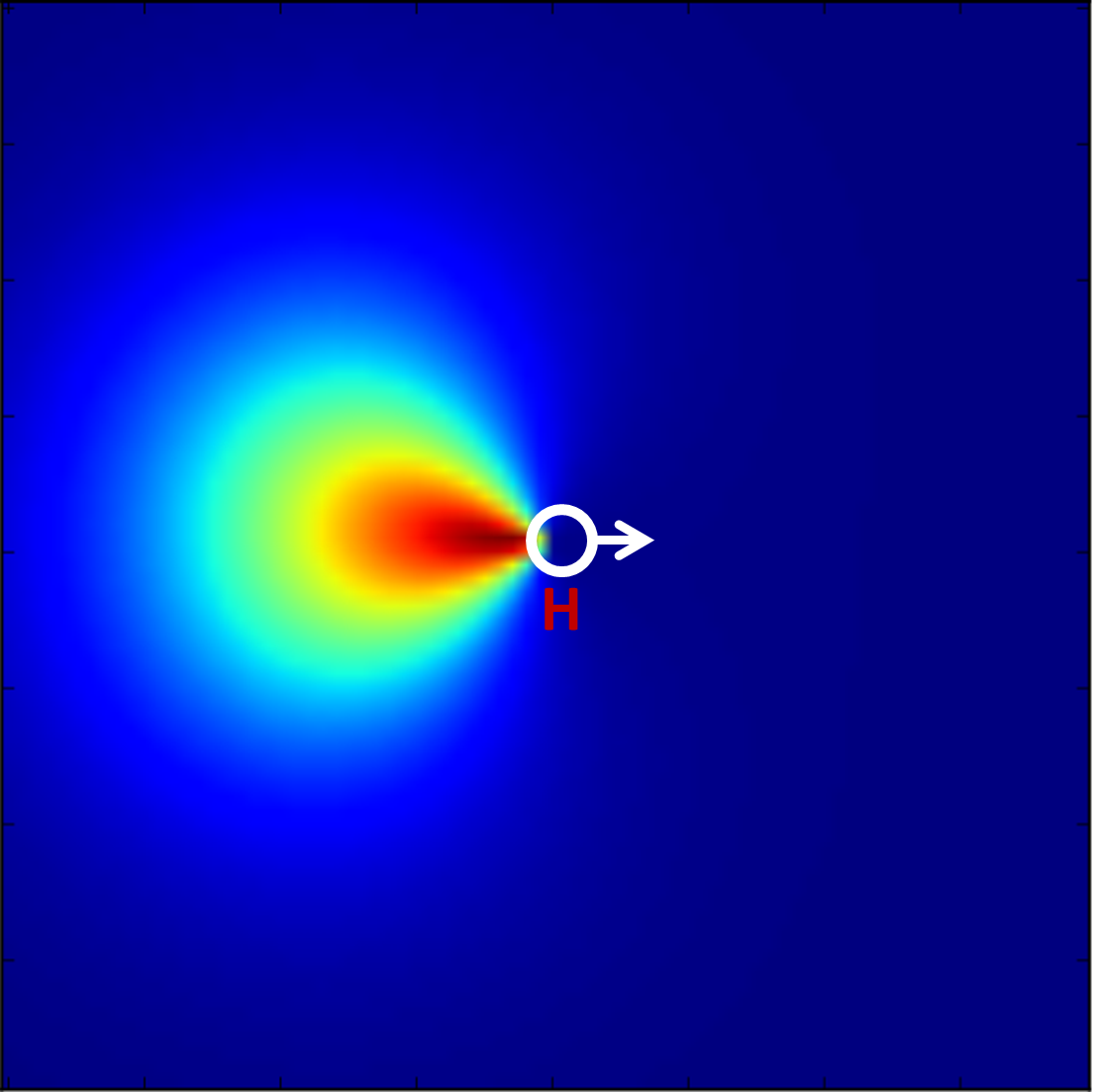}
}
\end{tabular}
\vspace*{\captionReduceTop}
\caption{\textbf{An example the learned planning affordance.} In the top-view, the human is at the center and facing the object
on the right (1m away). Dimension is 2m$\times$2m. (Best viewed in color)}
\vspace*{\captionReduceBot}
\vspace*{\captionReduceBot}
\label{fig:angular-pref-1}
\end{figure}

We consider the planning affordance for several activities (e.g., 
\textit{watching, interacting, working, sitting}, etc.). 
For each activity we model a separate cost function $\Psi_{a}$ and evaluate trajectories using Eq.~\eqref{eq:costfunction}. We consider two classes of activities:  in the first, the human and object are in
close proximity (\textit{sitting, working, reaching etc.}) and in the second, they
are at a distance (\textit{walking, watching, interacting etc.}). 
The affordance varies with the distance and angle between the human and the  object. We parameterize the cost as follows:
\begin{equation}
\Psi_a(t_i|E) = \begin{cases}
	      \Psi_{a,ang,h} \; \Psi_{a,ang,o} \; \Psi_{a,\beta} \\ \text{if }a \in \text{activities with human and}\\
	      \text{\;\;\;\;\;\;\;\;\;\;object at distance.}\\
	      \\
	      \Psi_{a,ang,h} \; \Psi_{a,dist,h} \\ \text{if } a\in
	      \text{activities with human and} \\
	      \text{\;\;\;\;\;\;\;\;\;\;object in close proximity.}
	      \end{cases}
\end{equation}

\begin{figure}[t]
\centering
\includegraphics[width=.6\linewidth,natwidth=401,natheight=312]{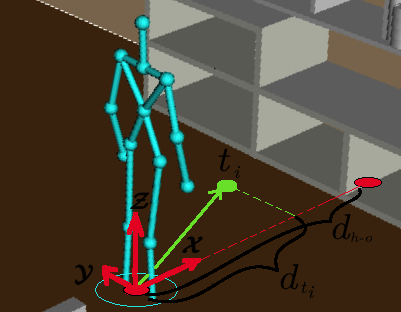}
\vspace*{.5\captionReduceTop}
\caption{\textbf{Human side of local co-ordinate system for watching activity.} 
Similar co-ordinates are defined for the object human interacts with. Unit vector $\mathbf{x}_{t_i}$ is 
the projection of waypoint $t_i$ on $x$-$y$ axis and normalized it by its length. Distance
between human and object is $d_{h-o}$, and $t_i$ projected on $x$-axis is of
length $d_{t_i}$.}
\vspace*{\captionReduceBot}
\vspace*{\captionReduceBot}
\label{fig:feature}
\end{figure}

\noindent \textit{Angular preference $\Psi_{a,ang}(\cdot)$:}  Certain angular positions w.r.t.\ the human and the object are more relevant for certain activities. For example, the spatial distribution for the \textit{watching} activity
is spread over a wider angle than the \textit{interacting} activity
(see Fig.~\ref{fig:angular-pref-1}). We capture the angular distribution of the activity in the space separating the human and the object with two cost functions  $\Psi_{a,ang,h}$ and $\Psi_{a,ang,o}$, centered at human and object respectively. For activities with close-proximity between the human and the object we define a single cost centered at human.
We parameterize the angular preference cost using the \textit{von-Mises} distribution as:
\begin{equation}
\label{eq:vonMisses}
\Psi_{a,ang,(\cdot)}(\mathbf{x}_{t_i};\mu,\kappa) = \frac{1}{2\pi
I_{0}(\kappa)}\exp(\kappa \mu^T \mathbf{x}_{t_i})
\end{equation}
In the above equation, $\mu$ and $\kappa$ are parameters that we will learn from the data, 
and $\mathbf{x}_{t_i}$ is a two-dimensional unit vector. As illustrated in Fig.~\ref{fig:feature}, we obtain $\mathbf{x}_{t_i}$ by 
projecting the waypoint $t_i$ onto the co-ordinate frame ($x$ and $y$ axis) defined locally  for the human-object activity. 

\smallskip
\noindent
\emph{Distance preference $\Psi_{a,dist}$:} The preferences vary with the robot distance from the human and the object. 
Humans do not prefer robots very close to them, 
especially when the robot is right-in-front or passes from behind~\cite{Sisbot07}. Fig.~\ref{fig:angular-pref-1}c shows the cost function learned by PlanIt. It illustrates that working humans do not prefer robots passing them from behind. We capture this by 
adding a 1D-Gaussian parameterized by a mean and variance, and centered at human. 
\\~
\emph{Edge preference $\Psi_{a,\beta}$:} 
For activities where the human and the object are separated by a distance, the preferences vary along the line connecting human and object. We parameterize this cost using a beta
distribution which captures the relevance of the activity along the human-object edge. Fig.~\ref{fig:edge-pref} illustrates that in the \textit{watching} activity users prefer robots to cross farther away
from them, whereas for the \textit{interacting} activity the
preference is symmetric w.r.t. the humans. To calculate this cost for the waypoint $t_i$, we first take its distance  from the human and project it along the line joining the human and the object $d_{t_i}$, and then normalize it by the distance $d_{h-o}$ between the human and the object. The normalized distance is
$\bar{d}_{t_i}=d_{t_i}/d_{h-o}$.
In the equation below, we learn the parameters $\alpha$ and $\beta$.\begin{equation}
\label{eq:beta}
\Psi_{a,\beta}(\bar{d}_{t_i};\alpha,\beta) = 
\frac{\bar{d}_{t_i}^{\alpha-1}(1-\bar{d}_{t_i})^{\beta-1}}{\text{B}(\alpha,\beta)}; \;\;\bar{d}_{t_i}\in
[0,1]
\end{equation}

The functions used in Eq.~\eqref{eq:costfunction} thus define our cost function.
This, however, has many parameters (30) that need to be learned from data.


\vspace*{\subsectionReduceTop}
\subsection{Generative Model: Learning the Parameters}
\vspace*{\subsectionReduceBot}
\label{subsec:learning}

\begin{figure}[t]
\center
\includegraphics[width=0.7\linewidth,natwidth=1517,natheight=876]{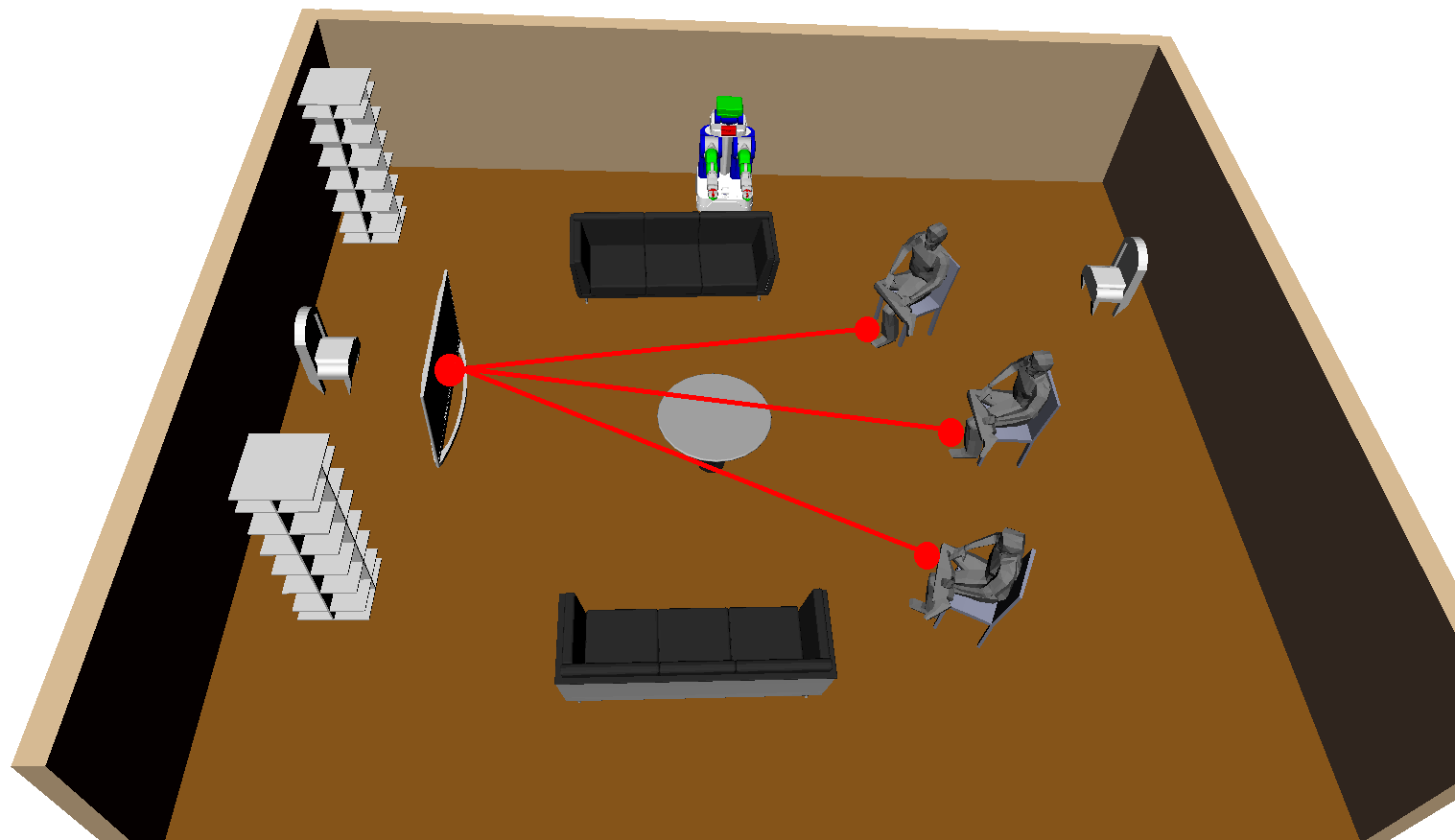}
\caption{\textbf{Watching activity.} Three humans watching a TV.}
\vspace*{\captionReduceBot}
\label{fig:generative-1}
\end{figure}

Given the user preference data from PlanIt we learn the parameters of Eq.~\eqref{eq:costfunction}. In order to keep the data collection easy we only elicit labels (bad, neutral or good) on the segments of the videos. The users do not reveal the human activity they think is being affected by the trajectory waypoint they labeled. In fact a waypoint can influence multiple activities. As illustrated in Fig.~\ref{fig:generative-1} a waypoint between the humans and the TV can affect multiple watching activities. 

\begin{figure}[t]
\center
\includegraphics[width=.7\linewidth,natwidth=780,natheight=284]{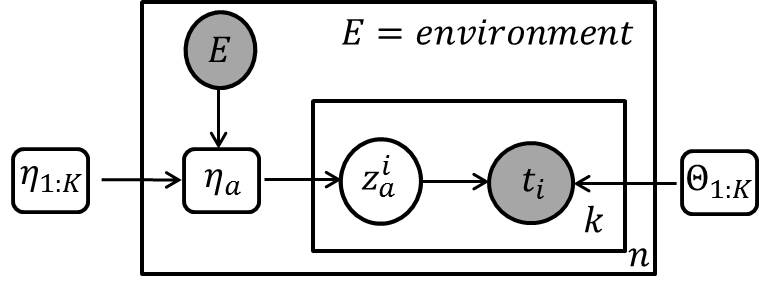}
\vspace*{\captionReduceTop}
\caption{\textbf{Feedback model.} Generative model of the user preference data.}
\vspace*{\captionReduceBot}
\vspace*{\captionReduceBot}
\label{fig:generative-2}
\end{figure}

We define a latent random variable $z_{a}^i \in \{0,1\}$ for the waypoint $t_i$; which is $1$ when the waypoint $t_i$ affects the activity $a$ and $0$ otherwise. From the user preference data we learn the following cost function:
\begin{equation}
\mathbf{\Psi}(\{t_1,..,t_k\}|E) = \prod_{i=1}^k \underbrace{\sum_{a \in
\mathcal{A}_E}
p(z_{a}^i|E) \; \Psi_{a}(t_i|E)}_\text{Marginalizing latent variable $z_a^i$}
\end{equation}
In the above equation, $p(z_a^i|E)$ (denoted with $\eta_a$) is the (prior) probability of user data arising from activity $a$, and $\mathcal{A}_E$ is the set of activities in environment $E$.\footnote{We extract
the information about the environment and activities by querying OpenRAVE. In practice
and in the robotic experiments, human activity information can be obtained using the
software package by Koppula et al.~\cite{Koppula13b}.} 
Fig.~\ref{fig:generative-2} shows the generative process
for preference data.

\smallskip
\noindent
\textit{Training data}: We obtain users preferences over $n$ environments
$E_1,..,E_n$. For each environment $E$ we consider $m$ trajectory segments $\mathcal{T}_{E,1},..,\mathcal{T}_{E,m}$ labeled as bad by users. For each segment $\mathcal{T}$ we sample $k$ waypoints
$\{t_{\mathcal{T},1},..,t_{\mathcal{T},k}\}$. We use $\Theta \in \mathbb{R}^{30}$ to denote the model parameters and solve the following maximum likelihood problem:
\begin{align}
\nonumber \Theta^* &=\arg\max_{\Theta} \prod_{i=1}^n \prod_{j=1}^m
\mathbf{\Psi}(\mathcal{T}_{E_i,j}|E_i;\Theta)\\
\label{eq:argmax} &= \begin{multlined}\arg\max_{\Theta} \prod_{i=1}^n \prod_{j=1}^m \prod_{l=1}^k
\sum_{a \in \mathcal{A}_{E_i}} \;  p(z_{a}^l|E_i;\Theta) \\   \Psi_{a}(t_{\mathcal{T}_{E_i,j},l}|E_i;\Theta)\end{multlined}
\end{align}
Eq.~\eqref{eq:argmax} does not have a closed form solution. We follow the
Expectation-Maximization (EM) procedure to learn the model parameters.
In the E-step we calculate the posterior activity assignment
$p(z_{a}^l|t_{\mathcal{T}_{E_i,j},l},E_i)$ for all the waypoints, and 
in the M-step we update the parameters. 

\smallskip
\noindent
\textbf{E-step}: In this step, with fixed model parameters, we calculate the posterior probability of an activity being affected by a waypoint, as follows:
\begin{equation}
p(z_{a}|t,E;\Theta) =
\frac{p(z_{a}|E;\Theta)\Psi_{a}(t|E;\Theta)}{\sum_{a \in \mathcal{A}_E}p(z_{a}|E;\Theta)\Psi_{a}(t|E;\Theta)}
\end{equation}
We calculate the above probability for every activity $a$ and the waypoint $t$ labeled by users in our data set. 

\smallskip
\noindent
\textbf{M-step}: Using the probabilities calculated in the E-step we update the model parameters in the M-step. Our affordance representation consists of three distributions, namely: Gaussian, von-Mises and
Beta. We update the parameters of the Gaussian, and the mean ($\mu$) of the von-Mises in closed form. To update the variance ($\kappa$) of the von-Mises we follow the first order approximation proposed by Sra~\cite{Sra12}.  Finally the parameters of the beta distribution ($\alpha$ and $\beta$) are updated approximately by using the first and the second order moments of the data. As an example below we give the M-step update of the mean $\mu_a$ of the von-Mises. 
\begin{equation}
\mu_a = \frac{\sum_{i=1}^n \sum_{j=1}^m \sum_{l=1}^k p(z_a^l |
\{t_{\mathcal{T}_{E_i,j},l}\},E_i)\mathbf{x}_{\{t_{\mathcal{T}_{E_i,j},l}\}}}{\|\sum_{i=1}^n \sum_{j=1}^m \sum_{l=1}^k p(z_a^l |
\{t_{\mathcal{T}_{E_i,j},l}\},E_i)\mathbf{x}_{\{t_{\mathcal{T}_{E_i,j},l}\}}\|}
\end{equation}
We provide the detailed derivation of the E and M-step in the supplementary material.\footnote{\url{http://planit.cs.cornell.edu/supplementary.pdf}}

\section{Experiments and Results}
\vspace*{\sectionReduceBot}
\label{sec:experiment}

Our data set consists of 122 context-rich 3D-environments that resemble real 
living rooms or bedrooms. We create them by downloading 2D-images of real 
environments from the Google images and reconstructing their corresponding 3D
models using OpenRAVE~\cite{Diankov10}.\footnote{For reconstructing 3D environments we download
3D object (.obj) files off the web, mainly from the Google warehouse.} We depict human activities by adding to the 3D models different human poses obtained
from the Kinect (refer Fig. 3 in~\cite{Jiang12} for the human poses).
In our experiments we consider six activities: \textit{walking, watching,
interacting, reaching, sitting} and \textit{working} as shown in 
Fig.~\ref{fig:activities}.

For these environments we generate trajectory videos and add them to the
PlanIt database. We crowdsource 2500 trajectory videos through PlanIt and for each trajectory a user labeled segments of it 
as \textit{bad, neutral} or \textit{good}, corresponding to the scores 
1, 3 and 5 respectively.
\begin{figure}[t]
\centering
\includegraphics[width=\linewidth,natwidth=950,natheight=675]{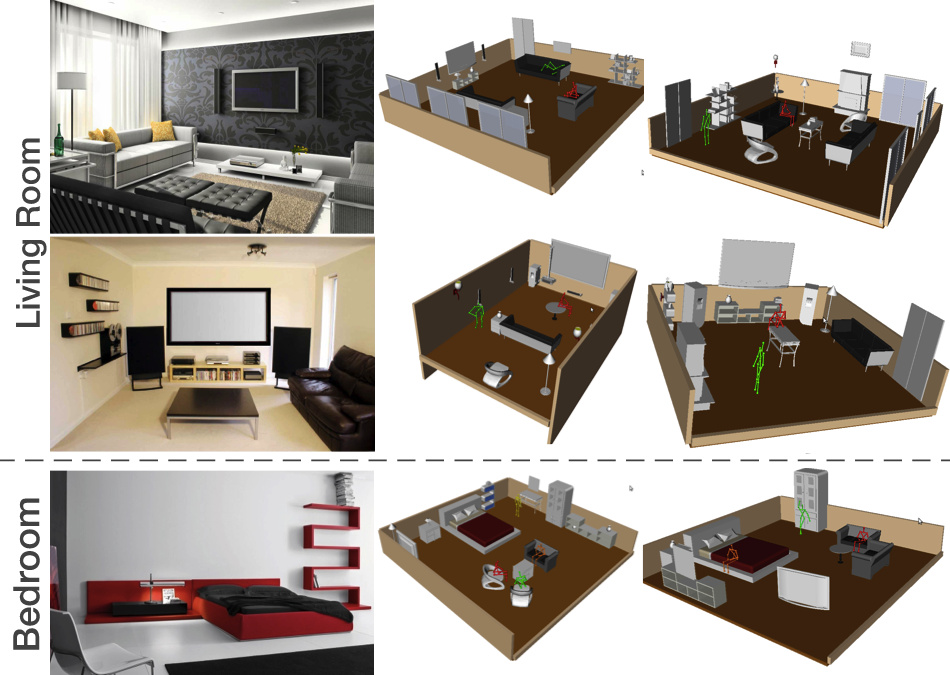}
\caption{\textbf{Examples from our dataset:} four living room  and two bedroom environments. 
On left is the 2D image we download from Google images. On right are the 3D reconstructed
environments in OpenRAVE.  All environments are rich in the types and number of objects and 
often have multiple  humans perform different activities. (Best view: Zoom and view in
color)}
\vspace*{\captionReduceBot}
\vspace*{\captionReduceBot}
\label{fig:environment}
\end{figure}

\subsection{Baseline algorithms}
\vspace*{\subsectionReduceBot}
We consider the following baseline cost functions:
\begin{itemize}[leftmargin=*]
\item \textit{Chance}: Uniformly randomly assigns a cost in interval [0,1] to a trajectory.
\item \textit{Maximum Clearance Planning (MCP)}: Inspired by Sisbot et
al.~\cite{Sisbot07}, this heuristic favors trajectories which stay farther away
from objects. The MCP cost of a trajectory is the (negated) 
root mean square distance from the nearest object across the trajectory waypoints. 
\item \textit{Human Interference Count (HIC)}: HIC cost of a trajectory is the
number of times it interferes with human activities. 
Interfering rules were hand designed 
on expert's opinion. 
\item \textit{Metropolis Criterion Costmap (MCC):} Similar to Mainprice et
al.~\cite{Mainprice11}, a trajectory's cost exponentially increases with
its closeness to surrounding objects. The MCC cost of a trajectory is defined as
follows: $c_{t_i} = \min_{o\in \mathcal{O}} dist(t_i,o)$
\begin{align*}
\Psi_{mc}(t_i) &= \begin{cases}
               		e^{-c_{t_i}}  & c_{t_i} < 1m\\
               		0 & \text{otherwise}
                     \end{cases}\\
MCC(\mathcal{T}=\{t_1,..,t_n\}) &= \frac{\sum_{i=1}^n \Psi_{mc}(t_i)}{n}
\end{align*} 
$dist(t_i,o)$ is the euclidean distance between the waypoint $t_i$ and the object $o$.
\item \textit{HIC scaled with MCC:} We design this heuristic by combining the HIC and the MCC costs. The HICMCC cost of a trajectory is $HICMCC(\mathcal{T}) = MCC(\mathcal{T})*HIC(\mathcal{T})$
\item \textit{Trajectory Preference Perceptron (TPP):} Jain et al.~\cite{Jain13}
learns a cost function from co-active user 
feedback in an online setting. We compare against the TPP using trajectory features from~\cite{Jain13}.
\end{itemize}

\noindent The above described baselines assign cost to trajectories and lower cost is preferred.
For quantitative evaluation each trajectory is also assigned a ground truth score based on the user feedback from PlanIt. The ground truth score of a trajectory is the minimum score given by a user.\footnote{The rationale behind this definition of the ground truth score is that a trajectory with a single bad waypoint is considered to be overall bad.} 
For example if two segments of a trajectory are labeled with scores 3 (neutral) and 5 (good), then the ground truth score is 3.  We denote the ground truth score of trajectory $\mathcal{T}$ as $score(\mathcal{T})$.

\subsection{Evaluation Metric} 
\vspace*{\subsectionReduceBot}
Given the ground truth scores we evaluate algorithms based on the following  metrics.

\begin{itemize}[leftmargin=*]
\setlength{\itemsep}{-1pt}
\setlength{\topsep}{-5pt}
\setlength{\parsep}{0pt}
\item \textit{Misclassification rate:} For a trajectory $\mathcal{T}_i$ we consider the set of trajectories $\mathbf{T}_i$ with higher ground truth score: 
$\mathbf{T}_i = \{\mathcal{T} | score(T) > score(\mathcal{T}_i)\}$.
The misclassification rate of an algorithm is the number of trajectories in $\mathbf{T}_i$ which it assigns a higher cost than $\mathcal{T}_i$.
We normalize this count by the number of trajectories in $\mathbf{T}_i$ and average it over all the trajectories $\mathcal{T}_i$ in the data set. Lower misclassification rate is desirable.
\item \textit{Normalized discounted cumulative gain (nDCG)~\cite{Manning08}:} This
metric quantifies how well an algorithm rank trajectories. It is a relevant metric
because autonomous robots can rank trajectories and execute the top ranked
trajectory~\cite{Dey12,Jain13}. Obtaining a rank list simply amounts to sorting the trajectories based on their costs. 
\end{itemize}

\subsection{Results}
\vspace*{\subsectionReduceBot}
We evaluate the trained model for its:
\begin{itemize}[leftmargin=*]
\item \textit{Discriminative power}: How well can the model distinguish good/bad trajectories?
\item \textit{Interpretability}: How well does the qualitative visualization of the cost function heatmaps match our intuition?
\end{itemize}

\subsection{Discriminative power of learned cost function}
\vspace*{\subsectionReduceBot}
\label{subsec:discriminative}

\begin{wrapfigure}{0}{.5\linewidth}
\centering
\vspace*{\captionReduceBot}
\includegraphics[width=\linewidth,natwidth=576,natheight=432]{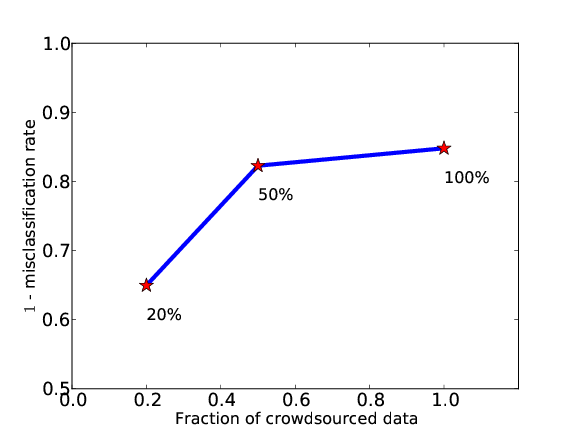}
\vspace*{\captionReduceBot}
\vspace*{\captionReduceBot}
\caption{\textbf{Crowdsourcing improves performance}: Misclassification
rate decreases as more users provide feedback
via PlanIt.}
\vspace*{.3\captionReduceBot}
\label{fig:crowd}
\end{wrapfigure}
A model trained on users preferences should reliably distinguish good
from bad trajectories i.e. if we evaluate two trajectories under the learned model then it should assign a lower cost to the better trajectory. We compare algorithms
under two training settings: (i) \textit{within-env}: we test and train on the
same category of environment using 5-fold cross validation, e.g. training on bedrooms 
and testing on new bedrooms; and (ii) \textit{cross-env}: we train on bedrooms
and test on living rooms, and vice-versa. In both settings the algorithms were
tested on environments not seen before.
\begin{table}
\centering
\caption{\textbf{Misclassification rate}: chances that an algorithm
presented with two trajectories (one good and other bad) orders them incorrectly.
Lower rate is better. The number inside bracket is standard error.}
\resizebox{\linewidth}{!}{
\begin{tabular}{r|c|c}\hline
{Algorithms}&{Bedroom}&{Living room}\\
\textit{Chance}&.52 (-) &.48 (-) \\
\textit{MCP based on Sisbot et al.~\cite{Sisbot07}}&.46 (.04)&.42 (.06)\\
\textit{MCC based on Mainprice et al.~\cite{Mainprice11}}&.44 (.03)&.42 (.06)\\
\textit{HIC}&.30 (.04)&.23 (.06)\\
\textit{HICMCC}&.32 (.04)&.29 (.05)\\
\textit{TPP based on Jain et al.~\cite{Jain13}}&.33 (.03)&.34 (.05)\\
\textit{Ours within scenario evaluation}&{.32 (.05)}&{.19 (.03)}\\\hline
\textit{Ours cross scenario evaluation}&\textbf{.27 (.04)}&\textbf{.17 (.05)}\\\hline
\end{tabular}
}
\vspace*{\captionReduceBot}
\label{tab:miss-rate}
\end{table}
\\~\\~
\begin{figure}[t]
\centering
\includegraphics[width=.5\linewidth,natwidth=576,natheight=432]{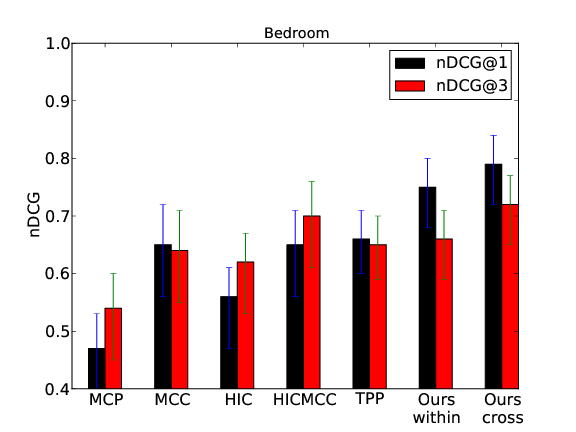}
\hskip -3mm
\includegraphics[width=.5\linewidth,natwidth=576,natheight=432]{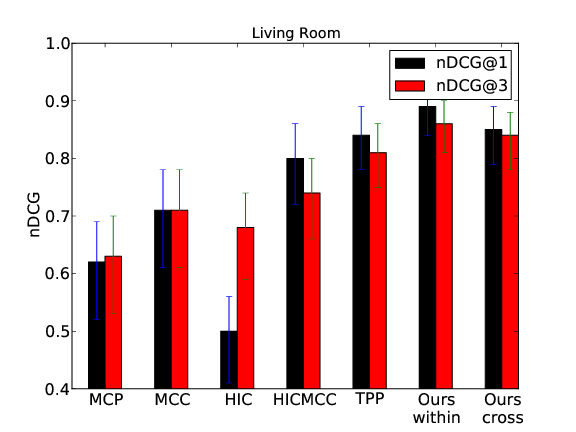}
\vspace*{\captionReduceTop}
\caption{\textbf{nDCG plots} comparing algorithms on bedroom (left) and living
room (right) environments. Error bar indicates standard error.}
\vspace*{\captionReduceBot}
\vspace*{\captionReduceBot}
\label{fig:ndcg_bar_chart}
\end{figure}
\textbf{How well algorithms discriminate between trajectories?}
We compare algorithms on the evaluation metrics described above.
As shown in Table~\ref{tab:miss-rate} our algorithm gives the lowest misclassification rate in comparison to the baseline algorithms. We also observe that the misclassification rate on bedrooms is lower with cross-env training than within-env. We conjecture this is because  of a harder learning problem (on average) when training on bedrooms than living rooms. In our data set, on average a bedroom have 3 human activities while a living room have 2.06. Therefore the model parameters
converge to better optima when trained on living rooms.  
As illustrated in Fig.~\ref{fig:ndcg_bar_chart}, our algorithm also ranks trajectories better than other baseline algorithms.
\\~\\~
\textbf{Crowdsourcing helps and so does learning preferences!} We compare our approach to TPP
learning algorithm by Jain et al.~\cite{Jain13}.  TPP learns with co-active
feedback which requires the user to iteratively improve the trajectory proposed
by the system. This feedback is time consuming to elicit and therefore difficult to
scale to many environments. On both evaluation metrics our crowdsourcing approach
outperforms TPP, which is trained on fewer environments.
Fig.~\ref{fig:crowd} shows our model improves as users provide more feedback
through PlanIt. Our data-driven
approach is also a better alternative to hand-designed cost functions.
\begin{figure}[t]
\centering
\includegraphics[width=.7\linewidth,natwidth=2538,natheight=1440]{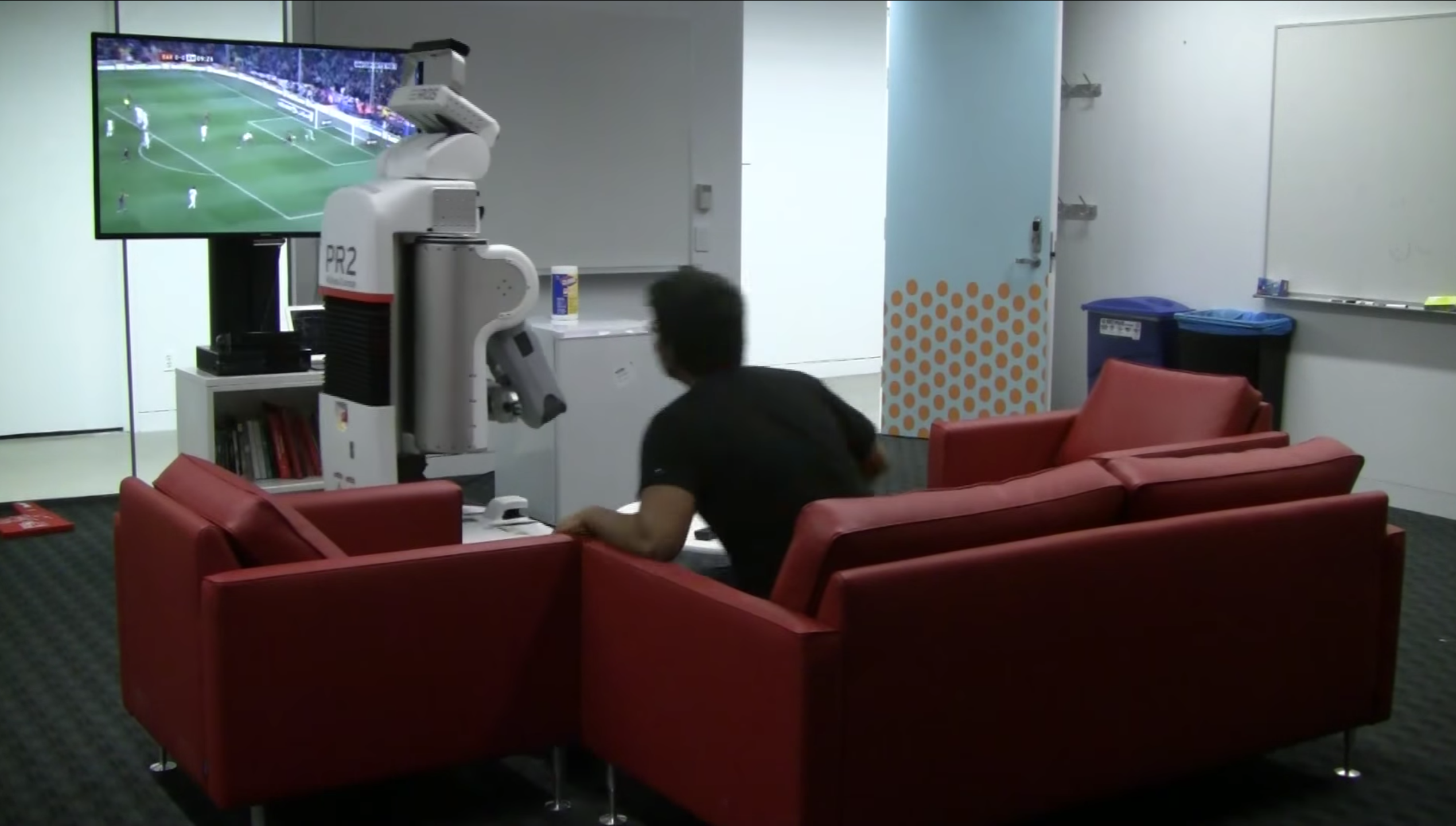}
\caption{\textbf{Robotic experiment:} A screen-shot of our algorithm running on
PR2. Without learning robot blocks view of human watching football. \url{http://planit.cs.cornell.edu/video}
}
\vspace*{\captionReduceBot}
\label{fig:robotexp}
\end{figure}

\begin{figure}[t]
\centering
\begin{tabular}{cccc}
\hskip -4mm
\subfigure[Edge preference for walking]{
\includegraphics[width=.4\linewidth,height=.3\linewidth,natwidth=576,natheight=432]{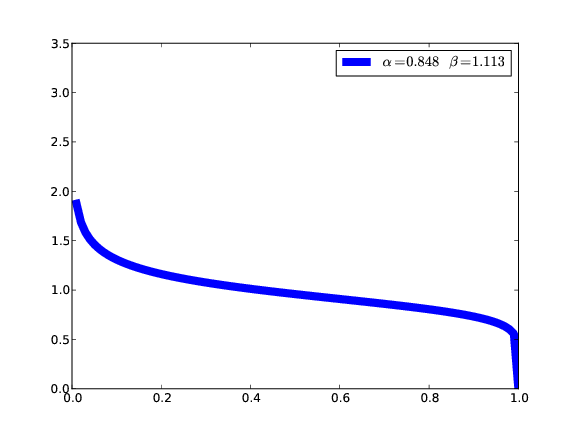}
}
&
\hskip -5mm
\subfigure[Walking]{
\includegraphics[width=.25\linewidth,height=.25\linewidth,natwidth=1131,natheight=1107]{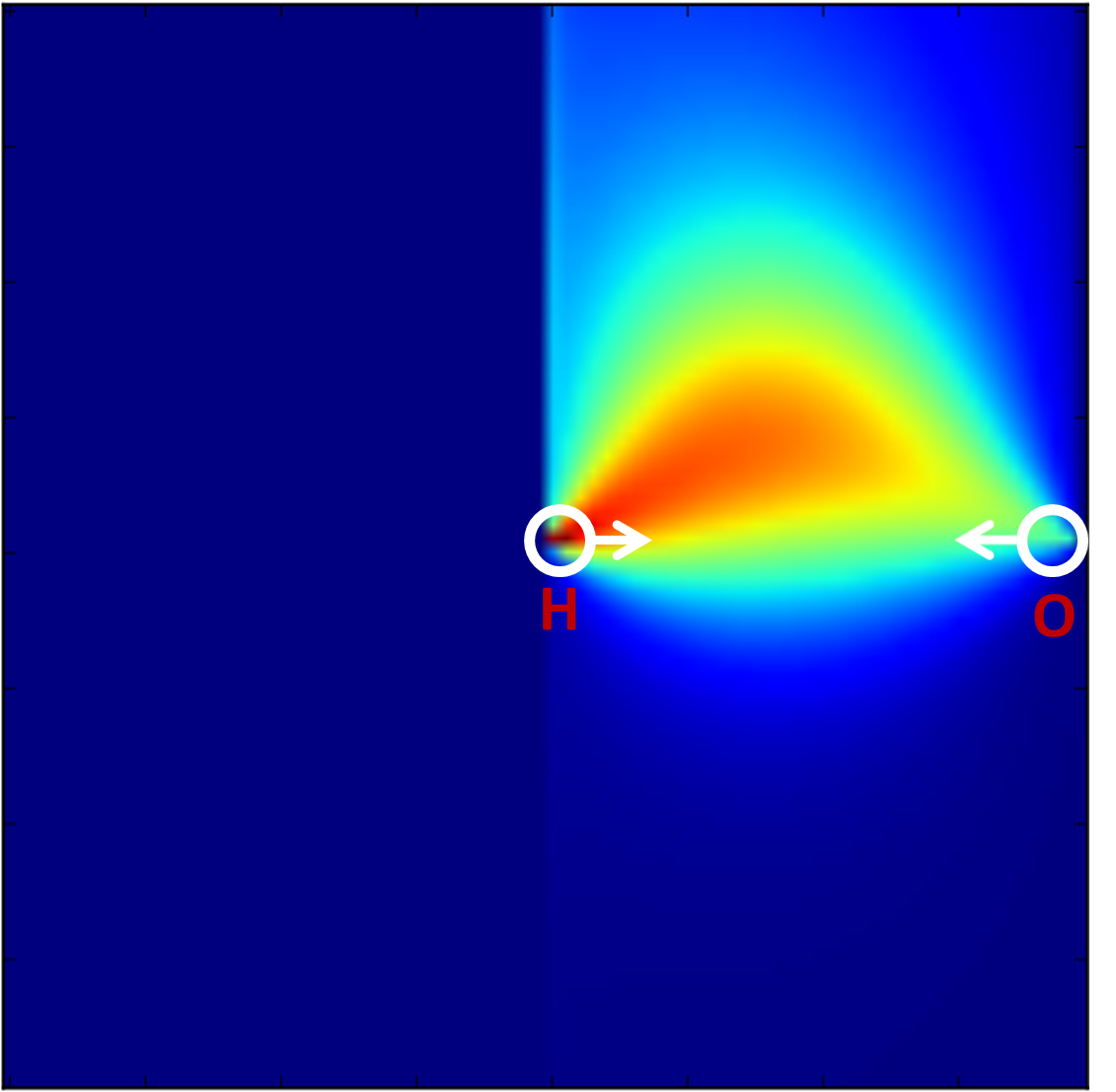}
}
&
\hskip -5mm
\subfigure[Sitting]{
\includegraphics[width=.25\linewidth,height=.25\linewidth,natwidth=1104,natheight=1105]{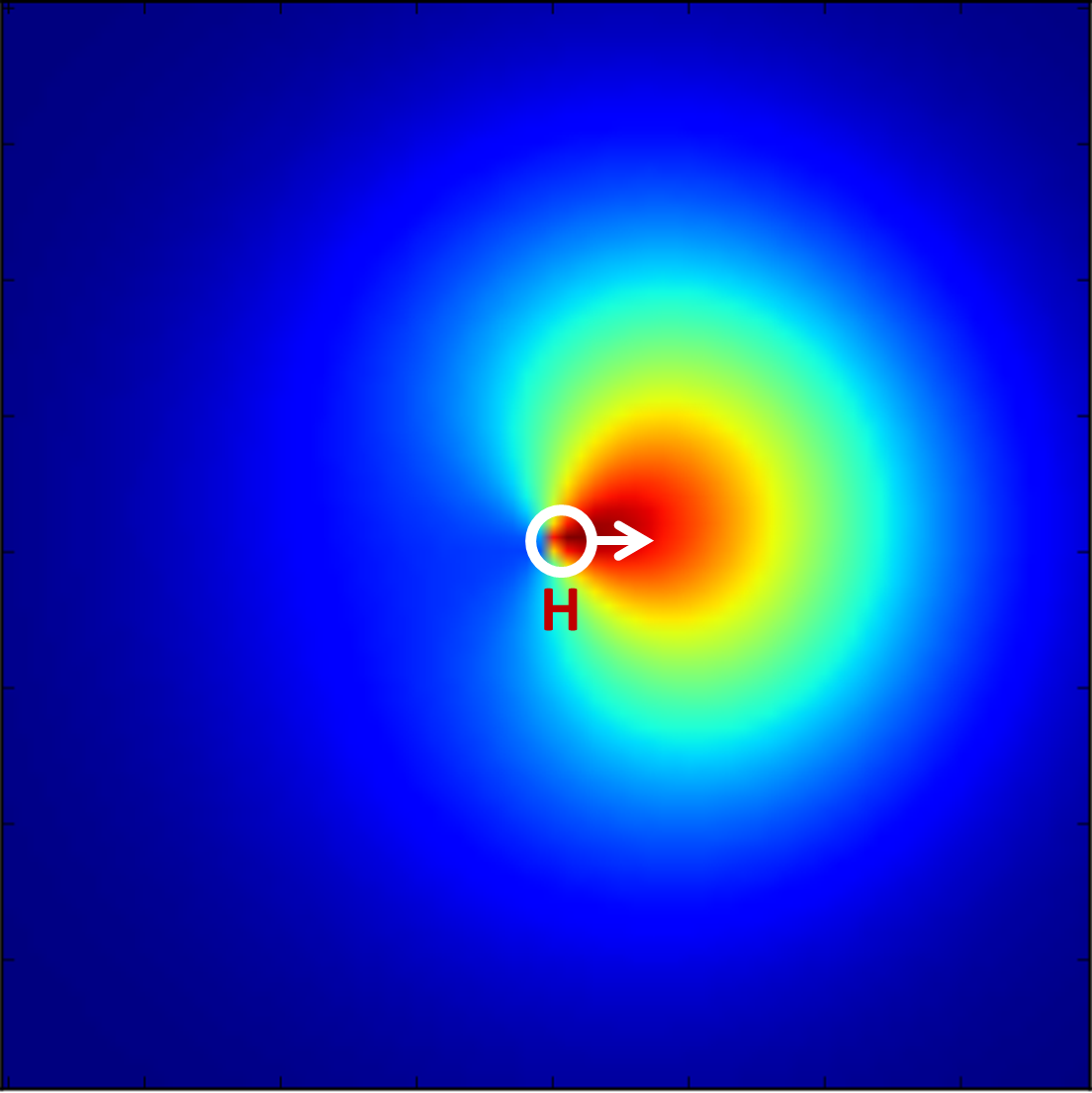}
}
\end{tabular}
\caption{\textbf{Learned affordance heatmaps}. (a) Edge preference for walking activity. Human is at 0 and
object at 1. (b,c) Top view of heatmaps. Human is at the center and facing right.}
\vspace*{\captionReduceBot}
\vspace*{\captionReduceBot}
\label{fig:remaining-act}
\end{figure}

\begin{figure}[t]
\centering
\begin{tabular}{cccc}
\subfigure[Bedroom]{
  \includegraphics[width=.45\linewidth,height=.3\linewidth,natwidth=640,natheight=480]{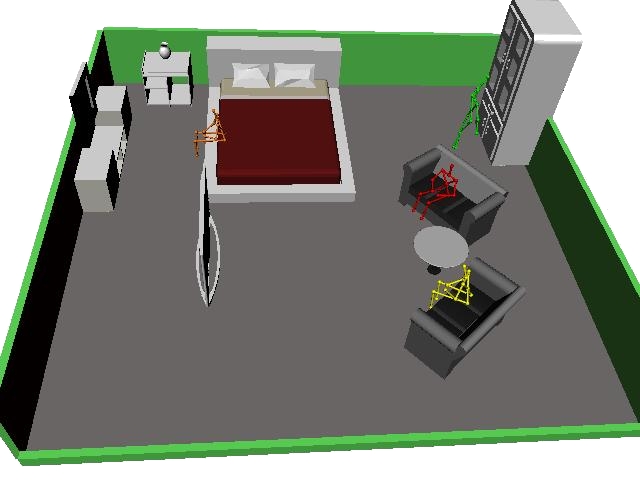}
}
&
\subfigure[Planning affordance]{
  \includegraphics[width=.45\linewidth,height=.3\linewidth,natwidth=562,natheight=450]{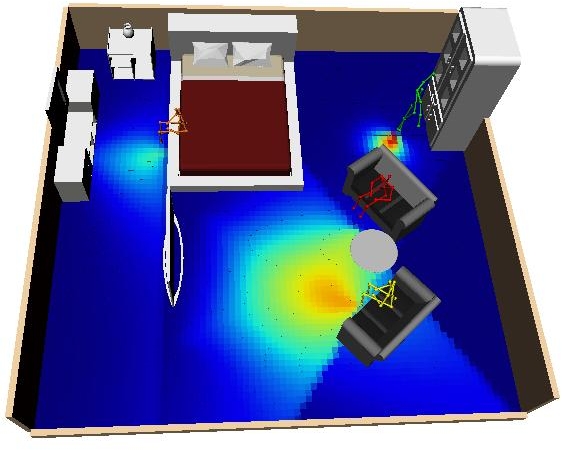}
}
\end{tabular}
\vspace*{\captionReduceTop}
\caption{\textbf{Planning affordance map for an environment}. Bedroom 
with sitting, reaching and watching activities. 
A robot uses the affordance map
as a cost for planning good trajectories.}
\vspace*{\captionReduceBot}
\label{fig:affordance-map}
\end{figure}
\subsection{Interpretability: Qualitative visualization of learned cost}
\vspace*{.5\subsectionReduceTop}
Visualizing the learned heatmaps is useful for understanding the spatial distribution of human-object activities. We discussed the learned heatmaps for watching, interacting and working activities in Section~\ref{sec:representation} (see Fig.~\ref{fig:angular-pref-1}).
Fig.~\ref{fig:remaining-act} illustrates the heatmap for the walking, and sitting activities.
\\~\\~
\textbf{How does crowd preferences change with the nature of activity?}
For the same distance between the human and the object, the spatial preference learned from the crowd is less-spread for interacting activity than watching and walking activities, as shown in Fig.~\ref{fig:angular-pref-1}. This empirical evidence implies that while interacting humans do not mind the robot in vicinity unless it blocks their eye contact.

The preferences also vary along the line joining the human and object. As shown in Fig.~\ref{fig:edge-pref}, while watching TV the space right in front of the human is critical, and the edge-preference rapidly decays as we move towards the TV. On the other hand, when human is walking towards an object the decay in
edge-preference is slower, Fig.~\ref{fig:remaining-act}(a). 
\\~\\~
Using the PlanIt system and the cost map of individual activities, a.k.a.\ affordance library, we generate the planning map of environments with multiple activities. Some examples of the planning map are shown
in Fig.~\ref{fig:affordance-map}.

\subsection{Robotic Experiment: Planning via PlanIt}
In order to plan trajectories in an unseen environment we generate its planning map and use it as an input cost to the RRT~\cite{Lavalle01} planner. Given the cost function, RRT plans trajectories with low cost. 
We implement our learned model on PR2 robot for the purpose of navigation when the human is watching a football match on the TV
(Fig.~\ref{fig:robotexp}). Without learning the preferences, the robot plans the shortest path to the goal and obstructs the human's view of TV. Demonstration: \url{http://planit.cs.cornell.edu/video}
\subsection{Application to manipulation tasks}
\vspace*{.5\subsectionReduceTop}
\label{sec:manip}
We also apply our model to manipulation tasks which requires
modeling higher dimensional state space involving 
object and robot-arm configurations. 
We consider tasks involving object-object interactions such as,
manipulating sharp objects in human vicinity or moving a glass of
water near electronic devices. For such tasks one
has to model both the object's distance from its surrounding and
its orientation.

\begin{figure}[t]
\center
\includegraphics[width=0.35\linewidth,natwidth=800,natheight=800]{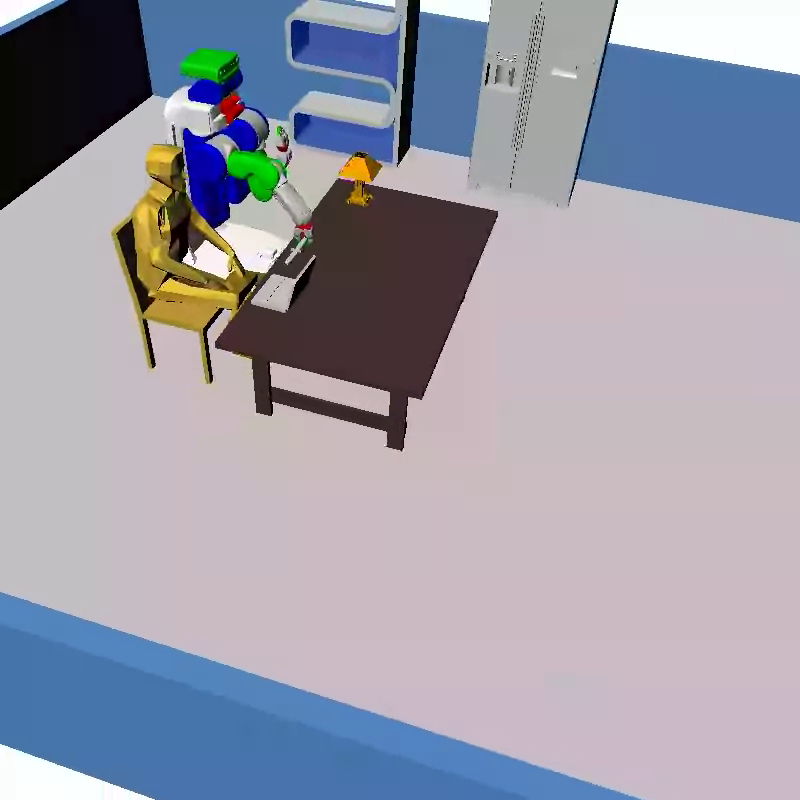}
\includegraphics[width=.25\linewidth,natwidth=207,natheight=300]{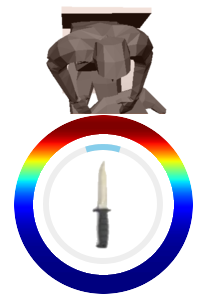}
\includegraphics[width=.35\linewidth,natwidth=640,natheight=610]{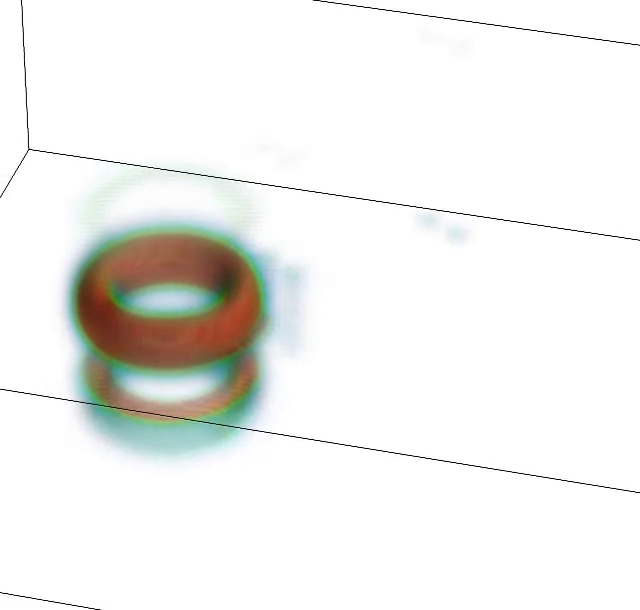}
\caption{
  \textbf{Learned heatmap for manipulation}.
  \textbf{(Left)} Robot manipulating knife in presence of human and laptop.
  \textbf{(Middle)} Learned heatmap showing preferred orientation of knife
w.r.t. human -- red being the most unpreferred orientation.
  \textbf{(Right)} 3D heat cloud of the same room showing undesirable
  positions of knife when its orientation is fixed to always point at the human.
  Heatmap has higher density near the face. (Best viewed in color. See
  interactive version at the PlanIt website.)
}
\vspace*{\captionReduceBot}
\vspace*{\captionReduceBot}
\label{fig:manip}
\end{figure}

Similar to navigation, we show videos on PlanIt of robot manipulating objects such as a
knife in the vicinity of humans and other objects. 
As feedback users label segments of videos as good/bad/neutral, we model the
feedback as a generative process shown in Fig.~\ref{fig:generative-2}.
We parametrize the cost function using Gaussian,
von-Mises and Beta distributions. More details on parameterization and EM updates
are in the supplementary
material.\footnote{\url{http://planit.cs.cornell.edu/supplementary.pdf}} 
Fig.~\ref{fig:manip} shows
learned preference for manipulating a knife in human vicinity. We learn that humans
prefer the knife pointing away from them. The learned heatmap is dense near 
the face implying humans strongly prefer the knife far away from their face.

\subsection{Sharing the learned concepts}
In order to make the knowledge from PlanIt useful to robots, we have partnered
with RoboBrain~\cite{Saxena14} -- an ongoing open-source research effort. The RoboBrain represents the information relevant for robots in form of a knowledge graph. The concepts learned by PlanIt are connected to its knowledge graph. The planning affordances learned by PlanIt are represented as nodes in the graph and they are connected (with edges) to their associated human activity node. This way PlanIt enables RoboBrain to connect different types of affordances associated with an activity.    
For example, the planning affordance when moving a knife and its grasping affordance~\cite{Koppula13} are related through \textit{cutting} activity. 

\section{Conclusion}
\vspace*{\sectionReduceBot}
In this paper we proposed a crowdsourcing approach for learning
user preferences over trajectories. Our PlanIt system is user-friendly and easy-to-use for eliciting large scale preference data from non-expert users. The simplicity of PlanIt comes at the expense of weak and noisy labels with latent annotator intentions. We presented a generative approach with latent nodes to model the preference data. Through PlanIt we learn spatial distribution of activities involving humans and objects. 
We experimented on many context-rich living and bedroom environments. Our results validate the advantages of crowdsourcing and learning the cost over hand encoded heuristics. We also implemented our learned model on the PR2 robot. PlanIt is publicly available for visualizing learned cost function heatmaps, viewing robotic demonstration, and providing preference feedback
\url{http://planit.cs.cornell.edu}.
\\~\\~
\textbf{Acknowledgement} This work was supported in part by ARO and by NSF Career Award to Saxena.

\bibliographystyle{IEEEtran}

\end{document}


\maketitle
\thispagestyle{empty}
\pagestyle{empty}

\section{Modified Planning Problem}\label{modified-planning-problem}

Unlike the 2-D navigation problem, manipulative tasks need to model the
3-D nature of the world and this problem depends on the object in hand
and the objects encountered while following a trajectory. For example,
bringing a knife close to any soft and fragile object is undesired. We
therefore take an object centric view to ground each object pair
interaction to a spatial distribution signifying object's functionality.
Hence the total cost definition at each waypoint is modified as:

\begin{equation}
  \boldsymbol{\Psi}(\mathcal{T} = \{t_1,...,t_n\})|E = \prod_j \prod_i \prod_{k} \Psi_{a_j,a_k}(t_i|E)
\end{equation}

Here we define attributes \(a_j\) and \(a_k\) respectively for grabbed
object and object in the vicinity of waypoint \(t_i\). These attributes are defined in the same way as in our previous work.\cite{c1}

\section{Extending the Generative
  Model}\label{extending-the-generative-model}

With the previous little change to the cost definition, the existing
\texttt{Plan-It} generative model can be extended to use the same user
preference data to model object-object attribute pair interaction
instead of environment activities and learn the 3-D spatial
distributions.

For each attribute pair, we define a different cost function
\(\Psi_{a_j, a_k}\). An environment can have multiple instances of
objects with same attributes and our overall cost function would have a
cumulative effect of different attribute pairs formed, while moving
through the waypoints.

\subsection{Modified Affordance
  Representation}\label{modified-affordance-representation}

The affordance representation gets slightly modified according to the
grabbed object attributes. For example for a knife:

\begin{equation}
  \Psi_{a_j, a_k}(t_i|E) = \begin{cases} \Psi_{a_j, a_k, dist} \Psi_{a_j,a_k, hei} \Psi_{a_j, a_k, ang} \quad \phantom{\infty} \text{if } a_k \in \text{human} \\ \Psi_{a_j, a_k, dist} \Psi_{a_j, a_k, ang} \quad \phantom{\infty} \text{if } a_k \notin \text{human} \end{cases}
\end{equation}

\emph{Distance Preference} \(\Psi_{a_j, a_k, dist}(.)\):
  Humans would not prefer objects like knives to get very close to them or
  any fragile object. That is the preferences vary with distance. This
  preference is captured in a 1-D Gaussian distribution centered around
  the object or human in the environment, parameterized by a mean and
  variance.
\emph{Angular Preference} \(\Psi_{a_j, a_k, ang}(.)\):
  Certain angular positions of the grabbed object w.r.t the human or
  object in the environment would be considered uncomfortable. For
  example, humans would not prefer knife pointed towards them, even if it
  is a reasonably far distance. This preference is captured by a
  \emph{von-Mises} distribution as:
  \[\Psi_{a_j, a_k, ang}(.) = \frac{1}{2\pi I_o(\kappa)} e^{\kappa \mu^T \boldsymbol{x}_{t_i}}\]
  In the above equation, \(\mu\) and \(\kappa\) are parameters that will
  be learned from the data, and \(x_{t_i}\) is a two dimensional unit
  vector representing the \(x\) and \(y\) projection of the angle between
  the grabbed-object orientation w.r.t to the object in the environment,
  where the coordinate system is defined locally for the attribute pair
  interaction.

\begin{figure}[htbp]
  \centering
  \includegraphics{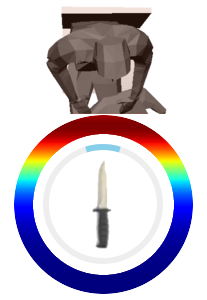}
  \caption{Relative Angle of Knife w.r.t sitting humans}
\end{figure}

\emph{Height Preference} \(\Psi_{a_j, a_k, hei}(.)\):
  It would not be preferable to move a \emph{sharp} knife over a pet.
  These preferences are captured by a beta distribution defined as:

  \begin{equation}
    \Psi_{a_j, a_k, hei}(.) = \frac{\bar{h}_{t_i}^{\alpha-1} (1 - \bar{h}_{t_i}^{\beta-1})}{\boldsymbol{B}(\alpha, \beta)}; \bar{h}_{t_i} \in [0,1]
  \end{equation}

  In the above equation, \(\bar{h_{t_i}}\) is defined as:

  \begin{equation}\bar{h}_{t_i} = \begin{cases} \frac{h_{t_i}}{h_{obj}} \quad \phantom{0} \text{if } \, \, h_{t_i} < h_{obj} \\ \frac{h_{max} - h_{t_i}}{h_{max}} \text{if } \, \, h_{t_i} > h_{obj}\end{cases}\end{equation}

  We learn the values of parameters \(\alpha\) and \(\beta\).

\subsection{Model}\label{model}

Extending the same activity based generative model to object-attribute
pairs, we get:

\begin{align}
  \Theta^{*} &= \arg \max_{\Theta} \prod_{x=1}^{n} \prod_{y}^{n} \boldsymbol{\Psi}{(\mathcal{T}_{E_x, y}|E_x;\Theta)} \\ &= \begin{multlined}\arg \max_{\Theta} \prod_{x=1}^{n} \prod_{y=1}^{m} \prod_{l=1}^{o} \sum_{k=1}^{|obj_{t_{\mathcal{T}_{E_x,y}, l}}|} p(z_{a_j, a_k} | E_x; \Theta) \\ \Psi_{a_j, a_k}(t_{\mathcal{T}_{E_x,y}, l}|E_x; \Theta)\end{multlined}
\end{align}

We use the same Expectation-Maximization (EM) approach to learn the
parameters. In the E-step, we calculate the posterior attribute pair
assignment \(p(z_{a_j, a_k}|t_{\mathcal{T}_{{E_x,y}}, l},E_x)\) for
every waypoint and use this to update the parameters in the M-step.

\textbf{E-step}: Keeping the model parameters fixed we find the
posterior probability of an attribute pair \(a_j, a_k\) at waypoint
\(t\):

\begin{equation}
  p(z_{a_j, a_k}|t, E;\Theta) = \frac{p(z_{a_j,a_k}|E;\Theta) \Psi_{a_j, a_k}(t|E;\Theta)}{\sum_{h=1}^{|obj|} p(z_{a_j, a_h})\Psi_{a_j, a_h}(t|E; \Theta)}
\end{equation}

\textbf{M-step}: Using the posterior from E-step we update the model
parameters. Our affordance representation consists of three
distributions: Gaussian, von-Mises and Beta. Gaussian parameters -- mean
(\(g\)) and variance (\(\sigma\)) and von-Mises mean (\(\mu\)) can be
updated in a closed form. We use Sra's \cite{c2} first order
approximation to update von-Mises variance (\(\kappa\)). We use a
similar approximation to update the beta distribution parameters
(\(\alpha\) and \(\beta\)) using the first and second order moments of
the data.

\textbf{Estimating Gaussian parameters}: For an attribute pair
\(a_j, a_k\):

\begin{equation}
  g_{a_j, a_k} = \frac{\sum_{x=1}^{n} \sum_{y=1}^{m} \sum_{l=1}^{o} p(z_{a_j, a_k}|t_{\mathcal{T}_{E_x, y},l}, E_x) d_{t_{\mathcal{T}_{E_x, y}, l}}}{\sum_{x=1}^{n} \sum_{y=1}^{m} \sum_{l=1}^{o} p(z_{a_j, a_k}|t_{\mathcal{T}_{E_x, y},l}, E_x)}
\end{equation}\begin{equation}
  \sigma_{a_j, a_k} = \frac{\sum_{x=1}^{n} \sum_{y=1}^{m} \sum_{l=1}^{o} p(z_{a_j, a_k}|t_{\mathcal{T}_{E_x, y},l}, E_x) (d_{t_{\mathcal{T}_{E_x, y}, l}} - g_{a_j, a_k})^2}{\sum_{x=1}^{n} \sum_{y=1}^{m} \sum_{l=1}^{o} p(z_{a_j, a_k}|t_{\mathcal{T}_{E_x, y},l}, E_x)}
\end{equation}

\textbf{Estimating Beta distribution parameters}: For an attribute pair
\(a_j, a_k\):

\begin{equation}
  m_{a_j, a_k} = \frac{\sum_{x=1}^{n} \sum_{y=1}^{m} \sum_{l=1}^{o} p(z_{a_j, a_k}|t_{\mathcal{T}_{E_x, y},l}, E_x) \bar{h}_{t_{\mathcal{T}_{E_x, y}, l}}}{\sum_{x=1}^{n} \sum_{y=1}^{m} \sum_{l=1}^{o} p(z_{a_j, a_k}|t_{\mathcal{T}_{E_x, y},l}, E_x)}
\end{equation}\begin{equation}
  v_{a_j, a_k} = \frac{\sum_{x=1}^{n} \sum_{y=1}^{m} \sum_{l=1}^{o} p(z_{a_j, a_k}|t_{\mathcal{T}_{E_x, y},l}, E_x) (\bar{h}_{t_{\mathcal{T}_{E_x, y}, l}} - g_{a_j, a_k})^2}{\sum_{x=1}^{n} \sum_{y=1}^{m} \sum_{l=1}^{o} p(z_{a_j, a_k}|t_{\mathcal{T}_{E_x, y},l}, E_x)}
\end{equation}

We now use these to estimate \(\alpha\) and \(\beta\):

\begin{equation}\alpha_{a_j, a_k} = m_{a_j, a_k} \left( \frac{m_{a_j, a_k}(1-m_{a_j, a_k})}{v_{a_j, a_k}} - 1\right)\end{equation}\begin{equation}\beta_{a_j, a_k} = (1-m_{a_j, a_k}) \left( \frac{m_{a_j, a_k}(1-m_{a_j, a_k})}{v_{a_j, a_k}} - 1\right)\end{equation}

\textbf{Estimating von-Mises distribution parameters}: For an attribute
pair \(a_j, a_k\):

\begin{equation}
  \mu_{a_j, a_k} = \frac{\sum_{x=1}^{n} \sum_{y=1}^{m} \sum_{l=1}^{o} p(z_{a_j, a_k}|t_{\mathcal{T}_{E_x, y},l}, E_x) \boldsymbol{x}_{\mathcal{T}_{E_x, y},l}}{||\sum_{x=1}^{n} \sum_{y=1}^{m} \sum_{l=1}^{o} p(z_{a_j, a_k}|t_{\mathcal{T}_{E_x, y},l}, E_x) \boldsymbol{x}_{\mathcal{T}_{E_x, y},l}||}
\end{equation}

To update \(\kappa\):

\begin{equation}\kappa_{a_j, a_k} = \frac{\bar{R}(2-\bar{R}^2)}{1 - \bar{R}^2}\end{equation}

\[\text{where, } \bar{R} = \frac{||\sum_{x=1}^{n} \sum_{y=1}^{m} \sum_{l=1}^{o} p(z_{a_j, a_k}|t_{\mathcal{T}_{E_x, y},l}, E_x) \boldsymbol{x}_{\mathcal{T}_{E_x, y},l}||}{\sum_{x=1}^{n} \sum_{y=1}^{m} \sum_{l=1}^{o} p(z_{a_j, a_k}|t_{\mathcal{T}_{E_x, y},l}, E_x)}\]

\textbf{Estimating hidden variable}: For an attribute pair \(a_j, a_k\):

\begin{equation}p(z_{a_j,a_k}|E;\Theta) = \frac{\sum_{x=1}^{n} \sum_{y=1}^{m} \sum_{l=1}^{o} p(z_{a_j, a_k}|t_{\mathcal{T}_{E_x, y},l}, E_x)}{N}\end{equation}

\[\text{where, } N = m \times n \times o\]

\addtolength{\textheight}{-12cm}   






\maketitle

\vspace*{\subsectionReduceTop}
\section{Generative Model: Learning the Parameters}
\vspace*{\subsectionReduceBot}
\label{subsec:learning}

\begin{figure}[t]
\center
\includegraphics[width=0.7\linewidth]{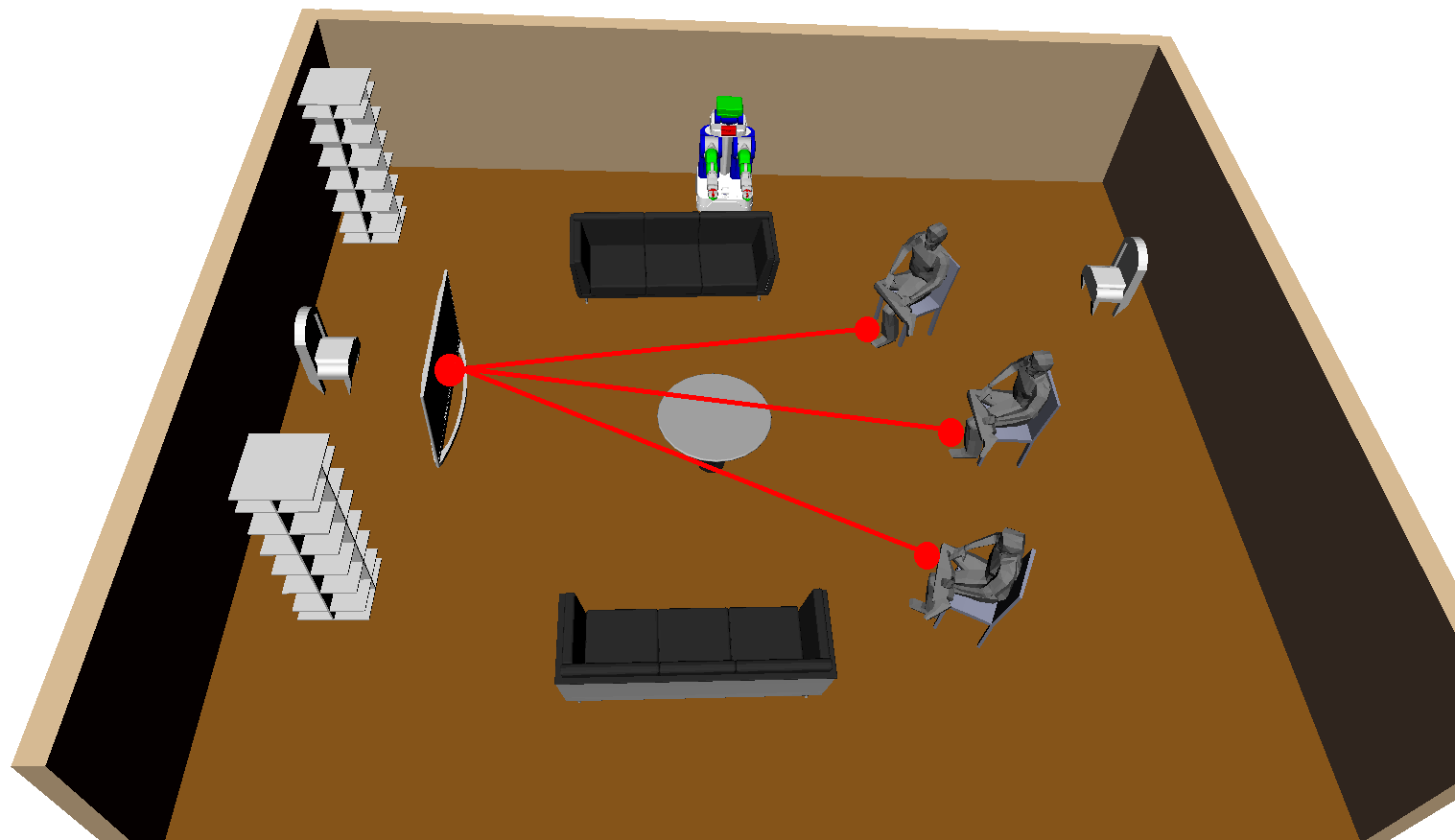}
\caption{An environment with three instances of watching activity.}
\vspace*{\captionReduceBot}
\label{fig:generative-1}
\end{figure}

Given user preference data from PlanIt, we 
learn the model parameters. 
Since our goal was to make the data collection easier for users, the labels we
get are either bad, neutral or good for a particular segment of the video.
The challenge is that we do not know which
activity $a$ is being affected by a given waypoint $t_i$ during feedback. 
A waypoint could even be influencing multiple activities. For example, in
Fig.~\ref{fig:generative-1} a waypoint passing between the human and TV could affect multiple watching activities. 

\begin{figure}[t]
\center
\includegraphics[width=.7\linewidth]{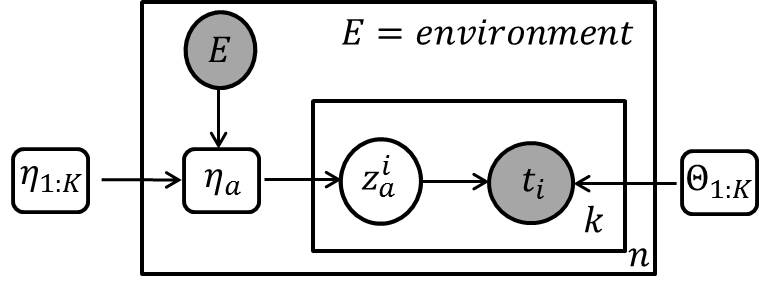}
\vspace*{\captionReduceTop}
\caption{Generative process for modeling the user preference data.}
\vspace*{\captionReduceBot}
\vspace*{\captionReduceBot}
\label{fig:generative-2}
\end{figure}

We therefore define a latent random variable $z_{a}^i \in \{0,1\}$ for waypoint $t_i$,
such that $p(z_a^i|E)$ (or $\eta_a$) is the (prior) probability of user data arising from activity
$a$.
Incorporating this parameter gives the following cost function:
\begin{equation}
\mathbf{\Psi}(\{t_1,..,t_k\}|E) = \prod_{i=1}^k \underbrace{\sum_{a \in
\mathcal{A}_E}
p(z_{a}^i|E) \; \Psi_{a}(t_i|E)}_\text{Marginalizing latent variable $z_a^i$}
\end{equation}

where $\mathcal{A}_E$ is the set of activities in environment $E$.\footnote{We extract
the information about the environment and activities by querying OpenRAVE. In practice
and in the robotic experiments, human activity information can be obtained using the
software package by Koppula et al.~\cite{Koppula13b}.} 
Figure~\ref{fig:generative-2} shows the generative process
for preference data.

\smallskip
\noindent
\textit{Training data}: We obtain user preferences over $n$ environments
$E_1,..,E_n$. For each environment $E$ we consider $m$ trajectory segments
$\mathcal{T}_{E,1},..,\mathcal{T}_{E,m}$ labeled as bad by users. For each 
segment $\mathcal{T}$ we sample $k$ waypoints
$\{t_{\mathcal{T},1},..,t_{\mathcal{T},k}\}$. We use $\Theta \in \mathbb{R}^{30}$ to denote 
the model parameters and solve the following maximum likelihood problem:
\begin{align}
\nonumber \Theta^* &=\arg\max_{\Theta} \prod_{i=1}^n \prod_{j=1}^m
\mathbf{\Psi}(\mathcal{T}_{E_i,j}|E_i;\Theta)\\
\label{eq:argmax} &=\arg\max_{\Theta} \prod_{i=1}^n \prod_{j=1}^m \prod_{l=1}^k
\sum_{a \in \mathcal{A}_{E_i}} \;  p(z_{a}^l|E_i;\Theta) \;   \Psi_{a}(t_{\mathcal{T}_{E_i,j},l}|E_i;\Theta)
\end{align}
Eq.~\eqref{eq:argmax} does not have a
closed form solution. We follow Expectation-Maximization (EM) procedure
to learn the model parameters. In E-step, we calculate the posterior activity assignment
$p(z_{a}^l|t_{\mathcal{T}_{E_i,j},l},E_i)$ for all the waypoints and 
update the parameters in the M-step. 

\smallskip
\noindent
\textbf{E-step}: In this step keeping the model parameters fixed we find the posterior
probability of a waypoint $t$ affecting an activity $a$.
\begin{equation}
p(z_{a}|t,E;\Theta) =
\frac{p(z_{a}|E;\Theta)\Psi_{a}(t|E;\Theta)}{\sum_{a \in \mathcal{A}_E}p(z_{a}|E;\Theta)\Psi_{a}(t|E;\Theta)}
\end{equation}
We calculate this posterior for every waypoint $t$ in our data. 

\smallskip
\noindent
\textbf{M-step}: Using the posterior from E-step we update the model parameters in this
step. Our affordance representation consists of three distributions, namely: Gaussian, von-Mises and
Beta. The parameters of Gaussian, and mean ($\mu$) of von-Mises are updated in a closed form. 
Following Sra~\cite{Sra12} we perform first order approximation to update the variance
($\kappa$) of von-Mises. The parameters of beta distribution ($\alpha$ and
$\beta$) are approximated using first and second order moments of the data.
\\~\\~
\textbf{Estimating von-Mises distribution parameters:} von-Mises is parameterized by a scalar
mean $\mu$ and variance $\kappa$. Mean for an activitiy $a$ has closed form update expression:
\begin{equation}
\mu_a = \frac{\sum_{i=1}^n \sum_{j=1}^m \sum_{l=1}^k p(z_a^l |
t_{\mathcal{T}_{E_i,j},l},E_i)\mathbf{x}_{t_{\mathcal{T}_{E_i,j},l}}}{\|\sum_{i=1}^n \sum_{j=1}^m \sum_{l=1}^k p(z_a^l |
t_{\mathcal{T}_{E_i,j},l},E_i)\mathbf{x}_{t_{\mathcal{T}_{E_i,j},l}}\|}
\end{equation}
However, updating $\kappa$ is not straightforward. We follow the first order
approximation by Sra~\cite{Sra12} and update $\kappa$ as follows:
\begin{align}
\kappa_a &= \frac{\bar{R}(2 - \bar{R}^2)}{1 - \bar{R}^2}\\
\text{where, } \bar{R} &= \frac{\|\sum_{i=1}^n \sum_{j=1}^m \sum_{l=1}^k p(z_a^l |
t_{\mathcal{T}_{E_i,j},l},E_i)\mathbf{x}_{t_{\mathcal{T}_{E_i,j},l}}\|}{\sum_{i=1}^n \sum_{j=1}^m \sum_{l=1}^k p(z_a^l |
t_{\mathcal{T}_{E_i,j},l},E_i)}
\end{align}
\\~
\\~
\textbf{Estimating Beta distribution parameters:} Beta distribution is
parameterized by two scalars $\alpha$ and $\beta$. We use method of moments to
estimate these parameters. For an activity $a$, we first estimate first and
second order moments i.e. sample mean and variance:
\begin{align}
m_a &= \frac{\sum_{i=1}^n \sum_{j=1}^m \sum_{l=1}^k p(z_a^l |
t_{\mathcal{T}_{E_i,j},l},E_i)\bar{d}_{t_{\mathcal{T}_{E_i,j},l}}}{\sum_{i=1}^n \sum_{j=1}^m \sum_{l=1}^k p(z_a^l |
t_{\mathcal{T}_{E_i,j},l},E_i)}\\
v_a &= \frac{\sum_{i=1}^n \sum_{j=1}^m \sum_{l=1}^k p(z_a^l |
t_{\mathcal{T}_{E_i,j},l},E_i)(\bar{d}_{t_{\mathcal{T}_{E_i,j},l}}-m_a)^2}{\sum_{i=1}^n \sum_{j=1}^m \sum_{l=1}^k p(z_a^l |
t_{\mathcal{T}_{E_i,j},l},E_i)}
\end{align}
We then estimate $\alpha$ and $\beta$ using the first and second order moments of
data:
\begin{align}
\alpha_a &= m_a\left( \frac{m_a(1-m_a)}{v_a} -1 \right)\\
\beta_a &= (1-m_a)\left( \frac{m_a(1-m_a)}{v_a} -1 \right)
\end{align}
\\~
\\~
\textbf{Estimating Gaussian distribution parameters:} It is parameterized by a
scalar mean $g$ and variance $\sigma$. For an activity $a$ we estimate parameters of Gaussian
distribution in closed form. 
\begin{align}
g_a &= \frac{\sum_{i=1}^n \sum_{j=1}^m \sum_{l=1}^k p(z_a^l |
t_{\mathcal{T}_{E_i,j},l},E_i){d}_{t_{\mathcal{T}_{E_i,j},l}}}{\sum_{i=1}^n \sum_{j=1}^m \sum_{l=1}^k p(z_a^l |
t_{\mathcal{T}_{E_i,j},l},E_i)}\\
\sigma_a &= \frac{\sum_{i=1}^n \sum_{j=1}^m \sum_{l=1}^k p(z_a^l |
t_{\mathcal{T}_{E_i,j},l},E_i)({d}_{t_{\mathcal{T}_{E_i,j},l}}-g_a)^2}{\sum_{i=1}^n \sum_{j=1}^m \sum_{l=1}^k p(z_a^l |
t_{\mathcal{T}_{E_i,j},l},E_i)}
\end{align}
In above equations ${d}_{t_{\mathcal{T}_{E_i,j},l}}$ is the distance of waypoint
$t_{\mathcal{T}_{E_i,j},l}$ from object/human.

\section{Application to Manipulation Scenarios}

Unlike the 2-D navigation problem, manipulative tasks need to model the
3-D nature of the world and this problem depends on the object in hand
and the objects encountered while following a trajectory. For example,
bringing a knife close to any soft and fragile object is undesired. We
therefore take an object centric view to ground each object pair
interaction to a spatial distribution signifying object's functionality.
Hence the total cost definition at each waypoint is modified as:

\begin{equation}
  \boldsymbol{\Psi}(\mathcal{T} = \{t_1,...,t_n\})|E = \prod_j \prod_i \prod_{k} \Psi_{a_j,a_k}(t_i|E)
\end{equation}

Here $a_j$ and $a_k$ are attributes of the grabbed object and objects in the
vicinity of waypoint $t_i$. These attributes are labels conveying physical
properties of the objects. For example, a knife can have an attribute ${sharp}$,
while a laptop can have attributes $electronic$ and $fragile$. These attributes
are defined similar to Jain et al.~\cite{Jain13}.
\subsection{Extending the Generative Model}

We now extend the 
\texttt{PlanIt} generative model to object-object attribute pair interaction
and learn the 3-D spatial
distributions.
For each attribute pair, we define a different cost function
\(\Psi_{a_j, a_k}\). An environment can have multiple instances of
objects with same attributes and our overall cost function would have a
cumulative effect of different attribute pairs formed, while moving
through the waypoints.

\subsubsection{Modified Affordance Representation:}

The affordance representation gets slightly modified according to the
grabbed object attributes. For example for a knife:

\begin{align}
  \Psi_{a_j, a_k}(t_i|E) = \begin{cases} \begin{multlined}\Psi_{a_j, a_k, dist} \Psi_{a_j,a_k, hei} \Psi_{a_j, a_k, ang} \\ \phantom{\infty} \text{if } a_k \in \text{human} \end{multlined} \\ \begin{multlined}\Psi_{a_j, a_k, dist} \Psi_{a_j, a_k, ang} \\ \phantom{\infty} \text{if } a_k \notin \text{human} \end{multlined} \end{cases}
\end{align}

\emph{Distance Preference} \(\Psi_{a_j, a_k, dist}(.)\):
Humans would not prefer objects like knives to get very close to them or
any fragile object i.e. the preferences vary with distance. This
preference is captured in a 1-D Gaussian distribution centered around
the object or human in the environment, parameterized by a mean and
variance.

\emph{Angular Preference} \(\Psi_{a_j, a_k, ang}(.)\):
Certain angular positions of the grabbed object w.r.t the human or
object in the environment would be considered uncomfortable. For
example, humans would not prefer knife pointed towards them, even if it
is a reasonably far distance. This preference is captured by a
\emph{von-Mises} distribution as:
\[\Psi_{a_j, a_k, ang}(.) = \frac{1}{2\pi I_o(\kappa)} e^{\kappa \mu^T \boldsymbol{x}_{t_i}}\]
In the above equation, \(\mu\) and \(\kappa\) are parameters that will
be learned from the data, and \(x_{t_i}\) is a two dimensional unit
vector representing the \(x\) and \(y\) projection of the angle between
the grabbed-object orientation w.r.t to the object in the environment,
where the coordinate system is defined locally for the attribute pair
interaction.

\begin{figure}[t]
  \center
  \includegraphics[width=0.30\linewidth,height=0.40\linewidth]{./rel-knife.png}
  \caption{Relative Angle of Knife w.r.t sitting humans}
\end{figure}

\emph{Height Preference} \(\Psi_{a_j, a_k, hei}(.)\):
It would not be preferable to move a \emph{sharp} knife over delicate objects or
humans.
These preferences are captured by a beta distribution defined as:

\begin{equation}
  \Psi_{a_j, a_k, hei}(.) = \frac{\bar{h}_{t_i}^{\alpha-1} (1 - \bar{h}_{t_i}^{\beta-1})}{\boldsymbol{B}(\alpha, \beta)}; \bar{h}_{t_i} \in [0,1]
\end{equation}

In the above equation, \(\bar{h_{t_i}}\) is defined as:

\begin{equation}\bar{h}_{t_i} = \begin{cases} \frac{h_{t_i}}{h_{obj}} \quad \phantom{0} \text{if } \, \, h_{t_i} < h_{obj} \\ \frac{h_{max} - h_{t_i}}{h_{max}} \text{if } \, \, h_{t_i} > h_{obj}\end{cases}\end{equation}

We learn the values of parameters \(\alpha\) and \(\beta\).

\subsubsection{Parameter Learning} We optimize the data likelihood:

\begin{align}
  \Theta^{*} &= \arg \max_{\Theta} \prod_{x=1}^{n} \prod_{y=1}^{m} \boldsymbol{\Psi}{(\mathcal{T}_{E_x, y}|E_x;\Theta)} \\ 
  &= \begin{multlined}\arg \max_{\Theta} \prod_{x=1}^{n} \prod_{y=1}^{m} \prod_{l=1}^{o} \sum_{k=1}^{|obj_{t_{\mathcal{T}_{E_x,y}, l}}|} p(z_{a_j, a_k} | E_x; \Theta) \\ \Psi_{a_j, a_k}(t_{\mathcal{T}_{E_x,y}, l}|E_x; \Theta)\end{multlined}
\end{align}

We use the Expectation-Maximization (EM) approach to learn the
parameters. In the E-step, we calculate the posterior attribute pair
assignment \(p(z_{a_j, a_k}|t_{\mathcal{T}_{{E_x,y}}, l},E_x)\) for
every waypoint and use this to update the parameters in the M-step.

\textbf{E-step}: Keeping the model parameters fixed we find the
posterior probability of an attribute pair \(a_j, a_k\) at waypoint
\(t\):

\begin{equation}
  p(z_{a_j, a_k}|t, E;\Theta) = \frac{p(z_{a_j,a_k}|E;\Theta) \Psi_{a_j, a_k}(t|E;\Theta)}{\sum_{h=1}^{|obj|} p(z_{a_j, a_h})\Psi_{a_j, a_h}(t|E; \Theta)}
\end{equation}

\textbf{M-step}: Using the posterior from E-step we update the model
parameters. Our affordance representation consists of three
distributions: Gaussian, von-Mises and Beta. Gaussian parameters -- mean
(\(g\)) and variance (\(\sigma\)) and von-Mises mean (\(\mu\)) can be
updated in a closed form. We use Sra's~\cite{Sra12} first order
approximation to update von-Mises variance (\(\kappa\)). We use a
similar approximation to update the beta distribution parameters
(\(\alpha\) and \(\beta\)) using the first and second order moments of
the data.

\textbf{Estimating Gaussian parameters}: For an attribute pair
\(a_j, a_k\):

\begin{equation}
  g_{a_j, a_k} = \frac{\sum_{x=1}^{n} \sum_{y=1}^{m} \sum_{l=1}^{o} p(z_{a_j, a_k}|t_{\mathcal{T}_{E_x, y},l}, E_x) d_{t_{\mathcal{T}_{E_x, y}, l}}}{\sum_{x=1}^{n} \sum_{y=1}^{m} \sum_{l=1}^{o} p(z_{a_j, a_k}|t_{\mathcal{T}_{E_x, y},l}, E_x)}
\end{equation}
\begin{align}
  \sigma_{a_j, a_k} &= \frac{\sum_{x=1}^{n} \sum_{y=1}^{m} \sum_{l=1}^{o} \begin{multlined}p(z_{a_j, a_k}|t_{\mathcal{T}_{E_x, y},l}, E_x) \\ (d_{t_{\mathcal{T}_{E_x, y}, l}} - g_{a_j, a_k})^2 \end{multlined}}{\sum_{x=1}^{n} \sum_{y=1}^{m} \sum_{l=1}^{o} p(z_{a_j, a_k}|t_{\mathcal{T}_{E_x, y},l}, E_x)}
\end{align}

\textbf{Estimating Beta distribution parameters}: For an attribute pair
\(a_j, a_k\):

\begin{equation}
  m_{a_j, a_k} = \frac{\sum_{x=1}^{n} \sum_{y=1}^{m} \sum_{l=1}^{o} p(z_{a_j, a_k}|t_{\mathcal{T}_{E_x, y},l}, E_x) \bar{h}_{t_{\mathcal{T}_{E_x, y}, l}}}{\sum_{x=1}^{n} \sum_{y=1}^{m} \sum_{l=1}^{o} p(z_{a_j, a_k}|t_{\mathcal{T}_{E_x, y},l}, E_x)}
\end{equation}
\begin{align}
  v_{a_j, a_k} = \frac{\sum_{x=1}^{n} \sum_{y=1}^{m} \sum_{l=1}^{o} \begin{multlined}p(z_{a_j, a_k}|t_{\mathcal{T}_{E_x, y},l}, E_x) \\ (\bar{h}_{t_{\mathcal{T}_{E_x, y}, l}} - g_{a_j, a_k})^2\end{multlined}}{\sum_{x=1}^{n} \sum_{y=1}^{m} \sum_{l=1}^{o} p(z_{a_j, a_k}|t_{\mathcal{T}_{E_x, y},l}, E_x)}
\end{align}

We now use these to estimate \(\alpha\) and \(\beta\):

\begin{equation}\alpha_{a_j, a_k} = m_{a_j, a_k} \left( \frac{m_{a_j, a_k}(1-m_{a_j, a_k})}{v_{a_j, a_k}} - 1\right)\end{equation}
\begin{equation}
  \beta_{a_j, a_k} = (1-m_{a_j, a_k}) \left( \frac{m_{a_j, a_k}(1-m_{a_j, a_k})}{v_{a_j, a_k}} - 1\right)
\end{equation}

\textbf{Estimating von-Mises distribution parameters}: For an attribute
pair \(a_j, a_k\):

\begin{equation}
  \mu_{a_j, a_k} = \frac{\sum_{x=1}^{n} \sum_{y=1}^{m} \sum_{l=1}^{o} p(z_{a_j, a_k}|t_{\mathcal{T}_{E_x, y},l}, E_x) \boldsymbol{x}_{\mathcal{T}_{E_x, y},l}}{||\sum_{x=1}^{n} \sum_{y=1}^{m} \sum_{l=1}^{o} p(z_{a_j, a_k}|t_{\mathcal{T}_{E_x, y},l}, E_x) \boldsymbol{x}_{\mathcal{T}_{E_x, y},l}||}
\end{equation}

To update \(\kappa\):

\begin{equation}\kappa_{a_j, a_k} = \frac{\bar{R}(2-\bar{R}^2)}{1 - \bar{R}^2}\end{equation}

\[\text{where, } \bar{R} = \frac{||\sum_{x=1}^{n} \sum_{y=1}^{m} \sum_{l=1}^{o} p(z_{a_j, a_k}|t_{\mathcal{T}_{E_x, y},l}, E_x) \boldsymbol{x}_{\mathcal{T}_{E_x, y},l}||}{\sum_{x=1}^{n} \sum_{y=1}^{m} \sum_{l=1}^{o} p(z_{a_j, a_k}|t_{\mathcal{T}_{E_x, y},l}, E_x)}\]

\textbf{Estimating hidden variable}: For an attribute pair \(a_j, a_k\):

\begin{equation}p(z_{a_j,a_k}|E;\Theta) = \frac{\sum_{x=1}^{n} \sum_{y=1}^{m} \sum_{l=1}^{o} p(z_{a_j, a_k}|t_{\mathcal{T}_{E_x, y},l}, E_x)}{N}\end{equation}

\[\text{where, } N = m \times n \times o\]





{\small
\bibliographystyle{IEEEtran}
\bibliography{COMTools/shortstrings,COMTools/references}
}
